\documentclass[10pt,twocolumn,letterpaper]{article}

\usepackage{cvpr}
\usepackage{times}
\usepackage{epsfig}
\usepackage{graphicx}
\usepackage{amsmath}
\usepackage{amssymb}
\usepackage{adjustbox}
\usepackage{algorithm}
\usepackage{multirow}

\newcommand{\squishlist}{
 \begin{list}{$\bullet$}
  { \setlength{\itemsep}{0pt}
     \setlength{\parsep}{3pt}
     \setlength{\topsep}{3pt}
     \setlength{\partopsep}{0pt}
     \setlength{\leftmargin}{1.5em}
     \setlength{\labelwidth}{1em}
     \setlength{\labelsep}{0.5em}}}

\newcommand{\squishlisttwo}{
 \begin{list}{$\bullet$}
  { \setlength{\itemsep}{0pt}
     \setlength{\parsep}{0pt}
    \setlength{\topsep}{0pt}
    \setlength{\partopsep}{0pt}
    \setlength{\leftmargin}{2em}
    \setlength{\labelwidth}{1.5em}
    \setlength{\labelsep}{0.5em} } }

\newcommand{\squishend}{
  \end{list}  }

\usepackage[pagebackref=true,breaklinks=true,letterpaper=true,colorlinks,bookmarks=false]{hyperref}

 \cvprfinalcopy 

\def\cvprPaperID{2062} 

\ifcvprfinal\fi
\begin{document}

\title{DeepHash: Getting Regularization, Depth and Fine-Tuning Right}


\renewcommand\footnotemark{}
\renewcommand\footnoterule{}

\author{
Jie Lin{$^{*,1,3}$},
Olivier Mor\`ere{$^{*,1,2,3}$}, 
Vijay Chandrasekhar{$^{1,3}$}, 
Antoine Veillard{$^{2,3}$}, 
Hanlin Goh{$^{1,3}$}%
\thanks{{$^{*}$}\, O. Mor\`ere and J. Lin contributed equally to this work.}
\thanks{1. Institute for Infocomm Research, A*STAR, Singapore.}
\thanks{2. Universit\'e Pierre et Marie Curie, Paris, France.}
\thanks{3. Image \& Pervasive Access Lab, UMI CNRS 2955, Singapore.} 
\thanks{We thank NVIDIA Corp. for donating the GPU used for this work.}\\
{I2R{$^{1}$}, UPMC{$^{2}$}, IPAL{$^{3}$}}}

\maketitle

\begin{abstract}


This work focuses on representing very high-dimensional global image descriptors using very compact 64-1024 bit binary hashes for instance retrieval. 
We propose DeepHash: a hashing scheme based on deep networks.
Key to making DeepHash work at extremely low bitrates are three important considerations -- regularization, depth and fine-tuning -- each requiring solutions specific to the hashing problem.
In-depth evaluation shows that our scheme consistently outperforms state-of-the-art methods across all data sets for both Fisher Vectors and Deep Convolutional Neural Network features, by up to 20$\%$ over other schemes. 
The retrieval performance with 256-bit hashes is close to that of the uncompressed floating point features -- a remarkable 512$\times$  compression.
\end{abstract}

\section{Introduction}

A compact binary image representation such as a 64-bit hash is a definite must for fast image retrieval.
64 bits provide more than enough capacity for any practical purposes, including internet-scale problems.
In addition, a 64-bit hash is directly addressable in RAM and enables fast matching using Hamming distances.

State-of-the-art global image descriptors such as Fisher Vectors (FV) \cite{Perronnin_CVPR_10} and Deep Convolutional Neural Network (DCNN) features \cite{AlexNet,Yandex} allow for robust image matching.
However, the dimensionality of such descriptors is typically very high: 8192 to 65536 floating point numbers for FVs\cite{Perronnin_CVPR_10} and 4096 for DCNNs \cite{AlexNet}.
Bringing such high-dimensional floating point representations down to a 64-bit hash is a considerable challenge.

Deep learning has achieved remarkable success in many visual tasks such as image classification~\cite{AlexNet,VeryDeepNeuralNets}, image retrieval~\cite{Yandex}, face recognition~\cite{deepface,deepid} and pose estimation~\cite{deeppose}. 
Furthermore, specific architectures such as stacked restricted Boltzmann machines (RBM) are primarily known as powerful dimensionality reduction techniques \cite{HintonRBM}.

We propose {\it DeepHash}, a deep binary hashing scheme that combines purpose-specific regularization with weakly-supervised fine-tuning.
A thorough empirical evaluation on a number of publicly available data sets shows that DeepHash consistently and significantly surpasses other state-of-the-art methods at bitrates from 1024 down to 64.
This is due to the correct mix of regularization, depth and fine-tuning.
This work represents a strong step towards the Holy Grail of a perfect 64-bit hash.

\begin{figure*}[ht] 
  \centering
    \includegraphics[width=0.98\hsize]{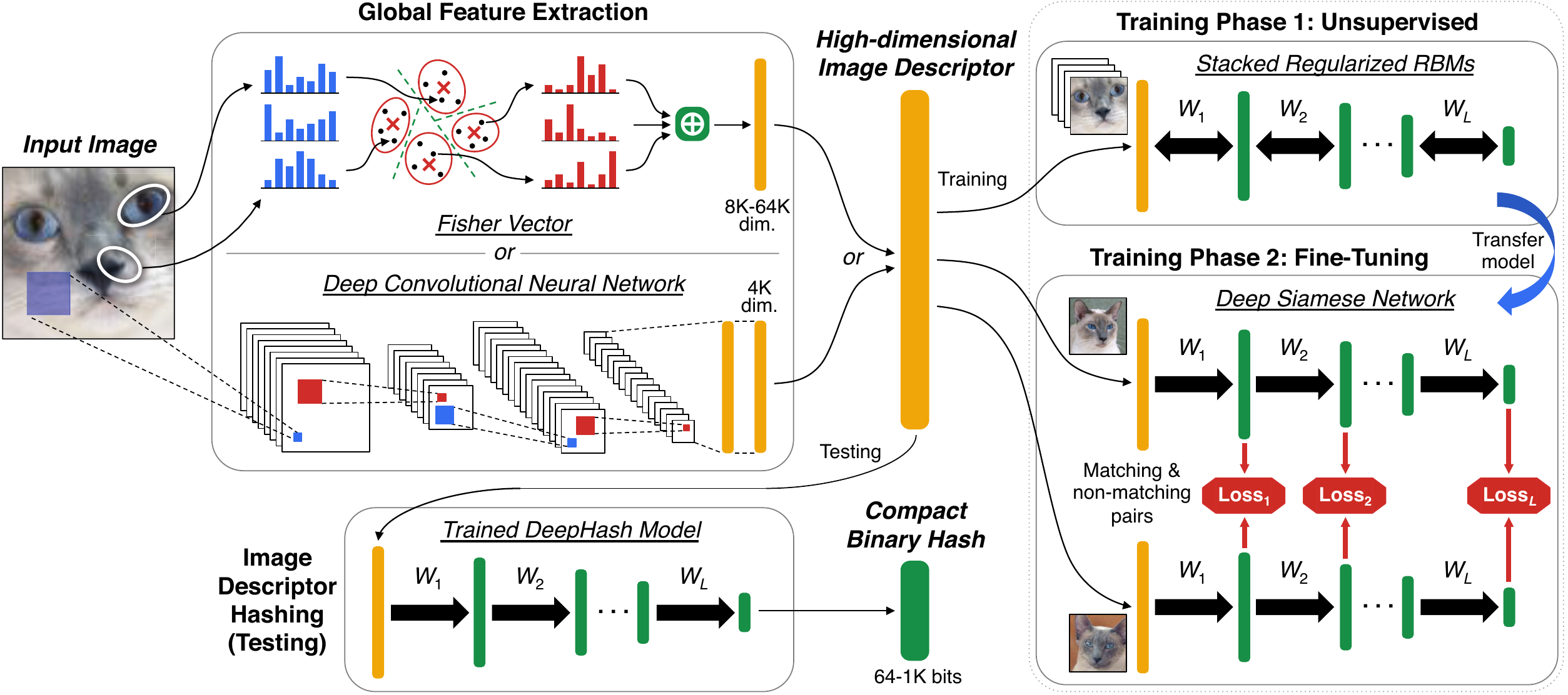}
  \caption{Our proposed hashing and model training pipeline. A high-dimensional global image descriptor, such as fisher vector and deep convolutional neural net feature, is extracted from an image. The trained DeepHash model transforms this image descriptor to a compact binary hash (between 64 to 1K bits), via a succession of $L$ nonlinear feedforward projections. The DeepHash model is trained in two phases: a unsupervised pre-training phase and a weakly supervised fine-tuning phase. In phase 1, restricted Boltzmann machines (RBMs) are trained in a layer-wise manner and stacked into a deep network. In phase 2, matching and non-matching pairs are used to construct deep Siamese networks for parameter fine-tuning.}
\label{fig:blockdiagram}
\end{figure*}

\section{Related Work and Contributions}

Hashing schemes can be broadly categorized into unsupervised and supervised (including semi-supervised) schemes. Examples of unsupervised schemes are Iterative Quantization~\cite{ITQ}, Spectral Hashing~\cite{SpectralHashing}, Restricted Boltzmann Machines~\cite{HintonRBM}, while some examples of state-of-the-art supervised schemes include Minimal Loss Hashing~\cite{MLH}, Kernel-based Supervised Hashing~\cite{KLSH}, Ranking-based Supervised Hashing~\cite{RSH} and Column Generation Hashing~\cite{CGH}.
Supervised hashing schemes are typically applied to the semantic retrieval problem. 
In this work, we are focused on instance retrieval: semantic retrieval is outside the scope of this work.

There is plenty of work on binary codes for descriptors like SIFT or GIST~\cite{ITQ,SemiSupervisedHashing,SphericalHashing,KristenHashingSurvey,SpectralHashing,KLSH,SKH,MLH,SmallCodes,SIFTSurvey,CHoG}.
There is comparatively little work on hashing descriptors like Fisher Vectors (FV) which are two orders of magnitude higher in dimensionality.
Perronnin et al.~\cite{Perronnin_CVPR_10} propose ternary quantization of FV, quantizing each dimension to +1,-1 or 0. 
Perronnin et al. also explore Locality Sensitive Hashing~\cite{RandomProjections} and Spectral Hashing~\cite{SpectralHashing}.
Spectral Hashing performs poorly at high rates, while LSH and simple ternary quantization need thousands of bits to achieve good performance.
Gong et al.  propose the popular Iterative Quantization (ITQ) scheme and apply it to GIST~\cite{ITQ}.
In subsequent work, Gong et al.~\cite{BPBC} focus on generating very long codes for global descriptors, and the Bilinear Projection-based Binary Codes (BPBC) scheme requires tens of thousands of bits to match the performance of the uncompressed global descriptor.
Jegou et al. propose Product Quantization (PQ) for obtaining compact representations~\cite{PQFisher}. 
While this produces compact descriptors, the resulting representation is not binary and cannot be compared with Hamming distances.
As opposed to previous work, our focus is on generating extremely compact binary representations for FV and DCNN features in the 64 bits-1024 bits range.

In this paper, we propose {\it DeepHash} (Figure~\ref{fig:blockdiagram}), a hashing scheme based on deep networks for high-dimensional global descriptors. The key to making the DeepHash scheme work at extremely low bitrates are three important considerations -- regularization, depth and fine-tuning -- each requiring solutions specific to the hashing problem.

\squishlist
\item
We pre-train a deep network using a RBM regularization scheme that is specifically adapted to the hashing problem. This enhances the efficiency of compact hashes, while achieving performance close to the uncompressed descriptor.

\item
Using stacked RBMs as a starting point, we fine-tune the model as a deep Siamese network. 
Critical improvements in the loss function lead to further improvements in retrieval results.

\item
DeepHash training is only required to be performed once on a single large independent training set. Through a thorough evaluation against state-of-the-art hashing schemes used for instance retrieval, we show that DeepHash outperforms other schemes by a significant margin of up 20$\%$, particularly at low bit rates. The results are consistently outstanding across a wide range of data sets and both DCNN and FV, showing the robustness of our scheme.
\squishend

\section{DeepHash}
\label{sec:deepHash}


{\it DeepHash} is a hashing scheme based on a deep network to generate binary compact hashes for image instance retrieval (Figure~\ref{fig:blockdiagram}).\footnote{DeepHash will be made publicly available on Caffe Model Zoo~\cite{Caffe} (\url{https://github.com/BVLC/caffe/wiki/Model-Zoo}).} Given a global image descriptor $\mathbf{z}^0$, a deep network performs a series of $L$ layers of nonlinear projections to generate a compact hash $\mathbf{z}^{L}$. The model is trained in two phases: 1) greedy layer-wise unsupervised pre-training with hashing regularization and 2) weakly-supervised Siamese fine-tuning.

In the unsupervised phase, stacked restricted Boltzmann machines (RBMs)~\cite{HintonDBN} are used to learn the initial parameters of the deep network. Each new layer in the network is trained to model the data distribution of the previous layer and is regularized specifically for hashing. A key feature is that this unsupervised pre-trained model is easily transferable. The unsupervised RBM parameters, which can be used to generate good hashes, can be further optimized with a fine-tuning phase. 
Fine-tuning is done through weak supervision by treating the deep model as a Siamese network~\cite{siamesenetwork}. 
Fine-tuning is also carried out an independent data set.
In the rest of this section, we will describe the details of the training process for our deep hashing scheme.

\subsection{Stacked Reguarized RBMs}
\label{sec:srbm}

The deep network with $L$ layers is initially pre-trained layer-by-layer from the bottom up through unsupervised learning, where each pair of successive layers ($\mathbf{z}^{l-1}$ and $\mathbf{z}^{l}$) is trained as an RBM building block. An RBM is an bipartite Markov random field with the input layer $\mathbf{z}^{l-1}\in\mathbb{R}^{I}$ connected to a latent layer $\mathbf{z}^{l}\in\mathbb{R}^{J}$ via a set of undirected weights $\mathbf{W}^{l}\in\mathbb{R}^{I J}$. The input units $z_{i}^{l-1}$ and latent units $z_{j}^{l}$ are also parameterised by their corresponding biases ${c}_{i}^{l-1}$ and $b_{j}^{l}$, respectively.

\paragraph{Binary RBMs.}
The first layer of the deep network takes a high-dimensional image descriptor as input. Previous works~\cite{Perronnin_CVPR_10,agrawal14analyzing} have shown that binarization of FV and DCNN features results in negligible loss in performance. For this work, binarization is done by component-wise mean thresholding for the inputs. 
We use binary latent units with sigmoid activation function, because binary output bits are desired for our hash. 
Binary RBMs are also faster and simpler to train as compared to continuous RBMs~\cite{RBMPracticalGuide}. 
All layers in the deep network will consist of binary units and binary hashes can be extracted from all intermediate layers.

The units within a layer are conditionally independent pairwise.
Therefore, the activation probabilities of one layer can be sampled by fixing the states of the other layer, and using distributions given by logistic functions for binary RBMs:
\begin{equation}
\mathbb{P}(z^{l}_j| \mathbf{z}^{l-1}) = 1/(1+\exp(-\mathbf{w}_{j} \mathbf{z}^{l-1} - b_j)),
\label{eq:alt_1}
\end{equation}
\begin{equation}
\mathbb{P}(z^{l-1}_i| \mathbf{z}^{l}) = 1/(1+\exp(-\mathbf{w}_{i}^{\top}\mathbf{z}^{l}-c_i)).
\label{eq:alt_2}
\end{equation}
As a result, alternating Gibbs sampling can be performed between the two layers. The sampled states are used to update the parameters $\{\mathbf{W}^{l},\mathbf{b}^{l},\mathbf{c}^{l-1}\}$ through minibatch gradient descent using the contrastive divergence algorithm~\cite{hintonCD} to approximate the maximum likelihood of the input distribution.

Given a trained RBM with fixed parameters and an input vector, a hash can be generated through a feedforward projection and thresholding Equation~(\ref{eq:alt_1}) at 0.5.
\begin{equation}
z^{l}_j=\begin{cases}
1,& \text{if } \mathbb{P}(z^{l-1}_i| \mathbf{z}^{l})>0.5\\
    0,              & \text{otherwise}.
\end{cases}
\label{eq:ffactivate}
\end{equation}

\paragraph{Hashing Regularization.}

The unsupervised RBM is naively trained without considering the task, which in this case is image hashing. It is, however, important for the RBMs to project the data in a latent subspace that is suitable for hashing. One way to encourage the learning of suitable representations is to perform regularization, such as sparsity~\cite{honglakSparsity,hintonSparsity,hanlinSparsity}. 
For classification, representations are encouraged to be very sparse to improve separability. 
For hashing, however, it is desirable to encourage the representation to make efficient use of the limited latent subspace.

For a given $l$ and a minibatch of input instances $\mathbf{z}_{\alpha}^{l-1}$, we add a regularization term to the RBM optimization problem to encourage (a) half the bits to be active for a given hash, and (b) each bit value to be equiprobable across hashes:
\begin{equation}\label{eq:sparseproblem}
\underset{\left\{\mathbf{W}^{l},\mathbf{b}^{l},\mathbf{c}^{l-1}\right\}}{\arg\min}\!-\!\sum_{\alpha}\log\!\bigg(\sum_{\mathbf{z}^{l}_{\alpha}\in\mathcal{E}_{\alpha}}\mathbb{P}(\mathbf{z}^{l-1}_{\alpha},\mathbf{z}^{l}_{\alpha})+\lambda h(\mathcal{E}_{\alpha})\!\bigg),\!
\end{equation}
where $\mathcal{E}_{\alpha}$ is the minibatch of sampled latent units for layer $l$ and $\lambda$ is the regularization constant.

We adapt the fine-grained regularization proposed in~\cite{hanlinSparsity} to suit our hashing problem. For each instance $\mathbf{z}^{l}_{\alpha}$, the regularization term for binary units penalises each unit $z_{j\alpha}^{l}$ with the cross entropy loss with respect to a target activation $t_{j\alpha}^{l}$ based on a predefined distribution,
\begin{equation}
h(\mathcal{E}_{\alpha})\!=\!-\!\!\!\!\sum_{\mathbf{z}^{l}_{\alpha}\in\mathcal{E}_{\alpha}}\!\!\!\sum_{j} t^{l}_{j\alpha}\log z^{l}_{j\alpha}\!+(1-t^{l}_{j\alpha})\log(1-z^{l}_{j\alpha}).\!\!
\end{equation}
Unlike~\cite{hanlinSparsity}, we choose the $t_{j\alpha}^{l}$ such that each $\{t_{j\alpha}^{l}\}_{j}$ for fixed $\alpha$ 
and each $\{t_{j\alpha}^{l}\}_{\alpha}$ for fixed $j$ is distributed according to $\mathcal{U}(0,1)$.
The uniform distribution is suitable for hashing high-dimensional vectors because the regularizer encourages the each latent unit to be active with a mean of $0.5$, while avoiding activation saturation. The result is a space-filling effect in the latent subspace, where data is efficiently represented.

After RBM training, we further enforce space utilization by substituting the learned RBM bias by the data set mean $\langle\mathbf{w}_{j} \mathbf{z}^{l-1}\rangle$ of the linear projection preceding the logistic. 
Equation~(\ref{eq:ffactivate}) is modified such that the final hash is centered around 0.5:
\begin{equation}
z^{l}_j=\begin{cases}
1,& \text{if } \,\mathbf{w}_{j} \mathbf{z}^{l-1}\!-\!\langle\mathbf{w}_{j} \mathbf{z}^{l-1}\rangle\!>\!0\\
    0,              & \text{otherwise}.
\end{cases}
\label{eq:biastrick}
\end{equation}

\paragraph{Stacked RBMs.}
The set of global image descriptors lie in a complex manifold in a very high-dimensional feature space. Deeper networks have the potential to discover more complex nonlinear hash functions and improve image instance retrieval performance. Following~\cite{HintonDBN}, we stack multiple RBMs by training one layer at a time to create a deep network with several layers.

Each layer models the activation distribution of the previous layer and captures higher order correlations between those units. 
For the hashing problem, we are interested in low-rate points of $64$, $256$ and $1024$ bits, which are typical operating points as discussed in Section~\ref{sec:exp}. 
We progressively decrease the dimensionality of latent layers by a factor of $2^{n}$ per layer, where $n$ is a tuneable parameter.
For our final models, $n$ is empirically selected for each layer resulting in variable network depth.





\begin{figure}
	\centering 
		\begin{tabular}{@{}c@{} @{}c@{}}
			\includegraphics[width=0.5\columnwidth]{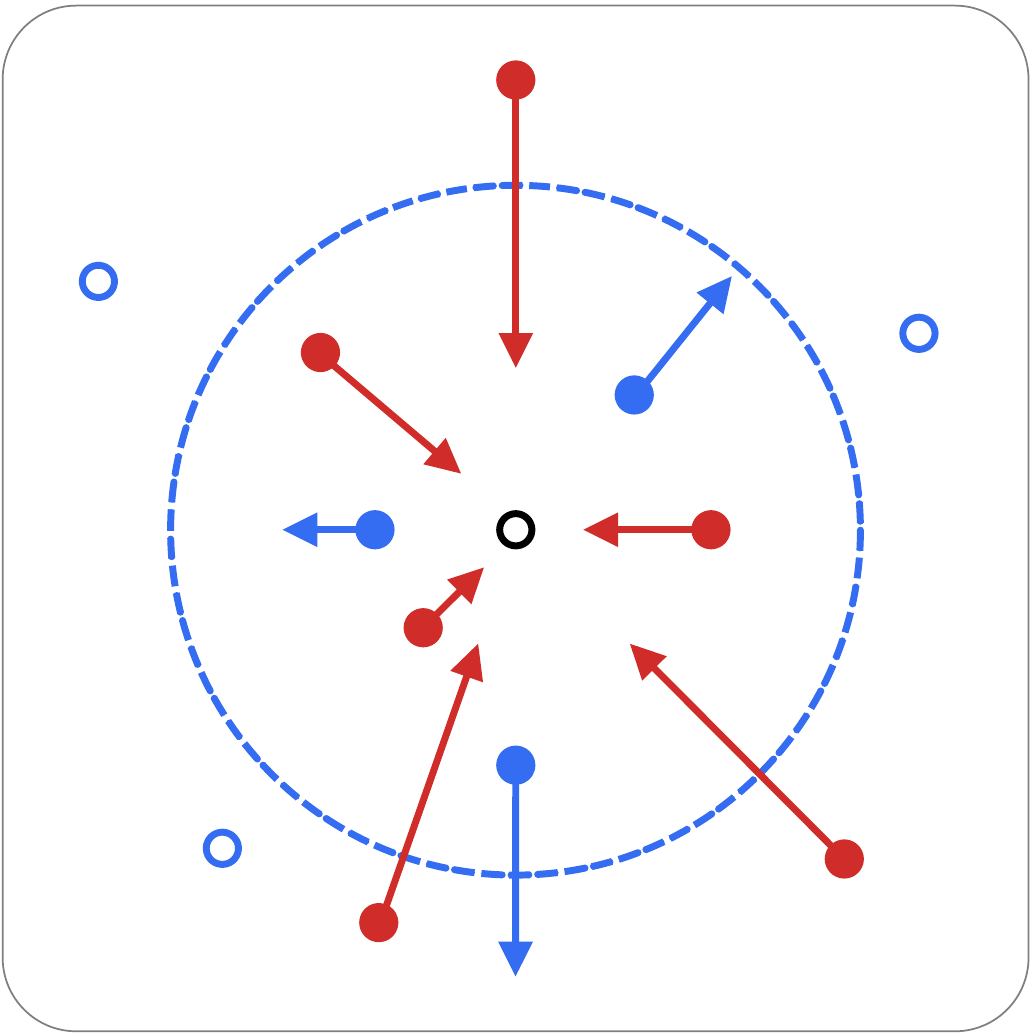} &
			\includegraphics[width=0.5\columnwidth]{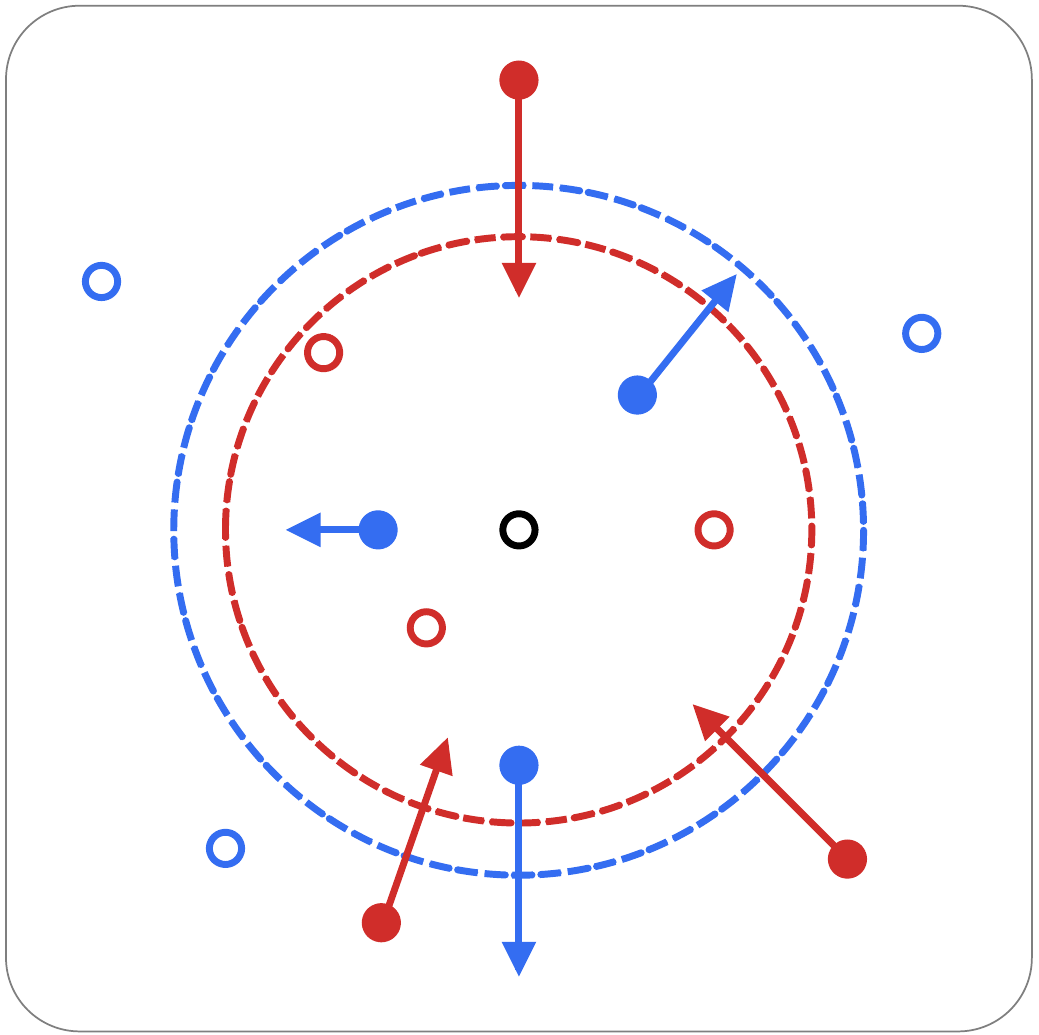} \\
			(a) & (b)
		\end{tabular}
		\caption{\footnotesize A sample point (black dot) with corresponding matching (red dots) and non-matching (blue dots) samples.
The contrastive loss used for fine-tuning can be interpreted as applying attractive forces between matching elements (red arrows) and repulsive forces between non-matching elements (blue arrows).
(a) The loss function (\ref{eq:siamese1}) proposed in \cite{Siamese} with a single margin parameter for non-matching pairs (blue circle). Matching elements are subject to attractive forces regardless of whether they are already close enough from each other which adversely affects fine-tuning.
(b) Our proposed loss function (\ref{eq:siamese2}) with an additional margin parameter affecting matching pairs reciprocally (red circle).}
	\label{fig:margin1}
\end{figure}

\subsection{Deep Siamese Fine-Tuning}

Retrieval results are driven by the structure of the local neighborhood around the query.
The unsupervised training is followed by a fine-tuning step in order to improve the local structure of the embedding.
The fine-tuning is performed with a learning architecture known as Siamese networks first introduced in \cite{siamesenetwork}.
The principle was later successfully applied to deep architectures for face identification \cite{chopra2005} and shown to produce representations robust to various transformations in the input space \cite{Siamese}.
The use of Siamese architectures in the context of image retrieval from DCNN features was recently suggested as a possible improvement to the state-of-the-art on the subject \cite{Yandex}. 

A Siamese network is a weakly-supervised scheme for learning a similarity measure from pairs of data instances labeled as matching or non-matching.
In our adaptation of the concept, the weights of the trained RBM network are fine-tuned by learning a similarity measure at every intermediate layer in addition to the target space.
Given a pair of data $(\mathbf{z}_\alpha^0,\mathbf{z}_\beta^0)$, a contrastive loss $\mathcal{D}_l$ is defined for every layer $l$ and the error is back propagated though gradient descent.
Back propagation for the losses of individual layers ($l = 1..L$) is performed at the same time.
Applying the loss function proposed by Handsell et al.~in \cite{Siamese} yields:
\begin{equation}
\label{eq:siamese1}
\mathcal{D}_l(\mathbf{z}_\alpha^0,\mathbf{z}_\beta^0) = y \lVert \mathbf{z}_\alpha^l - \mathbf{z}_\beta^l \rVert_2^2 + (1-y) \max(m - \lVert \mathbf{z}_\alpha^l - \mathbf{z}_\beta^l \rVert_2^2, 0)
\end{equation}
where $y=1$ if $(\mathbf{z}_\alpha^0,\mathbf{z}_\beta^0)$ is a matching pair or $y=0$ otherwise, and $m > 0$ is a margin parameter affecting non-matching pairs.
As shown in Figure~\ref{fig:margin1}(a), the effect is to apply a contractive force between elements of any matching pairs and a repulsive force between elements of non-matching pairs which element-wise distance is shorter than $\sqrt{m}$.

\begin{figure}
	\centering
		\includegraphics[width=\columnwidth]{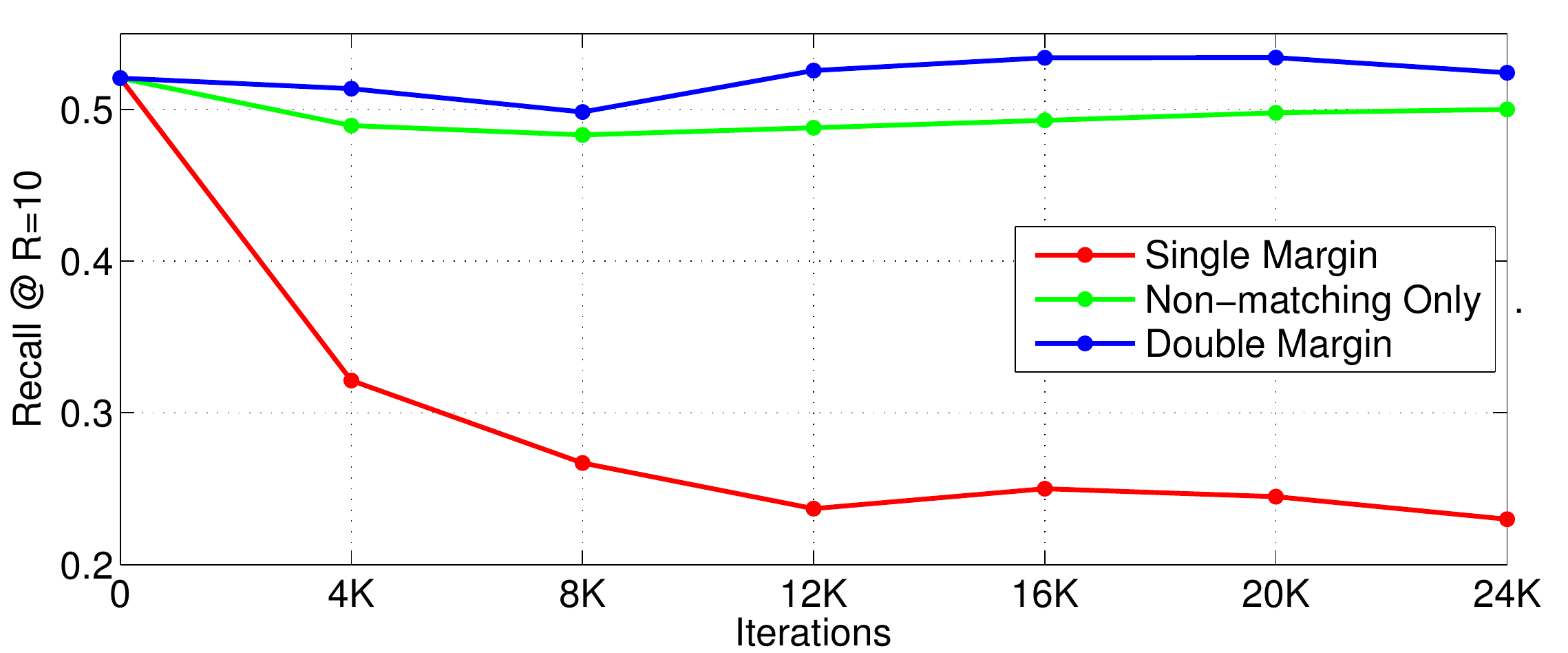}
		\caption{\footnotesize Recall @ R=10 on the {\it Holidays} data set (See Section~\ref{sec:eval_framework} for a description of the data sets) over several iterations of Siamese fine-tuning.
The recall rate quickly collapses when using the single margin loss function suggested by Hadsell et al. \cite{Siamese} while performance is better retained when only non-matching pairs are passed.
The double-margin loss solves the problem.
The network is a stacked RBM (8192-4096-2048-64) trained with Fisher descriptors on the {\it ImageNet} data set. Matching pairs are sampled from the {\it Yandex} data set. 
For every matching pair, a random non-matching element is chosen from the data set to form two non-matching pairs.
There are 33 matching pairs and 66 corresponding non-matching pair with every iteration.
The test set is the Holidays data set. 
}
	
	\label{fig:margin3}
\end{figure}

\begin{figure}
	\centering
		\includegraphics[width=\columnwidth]{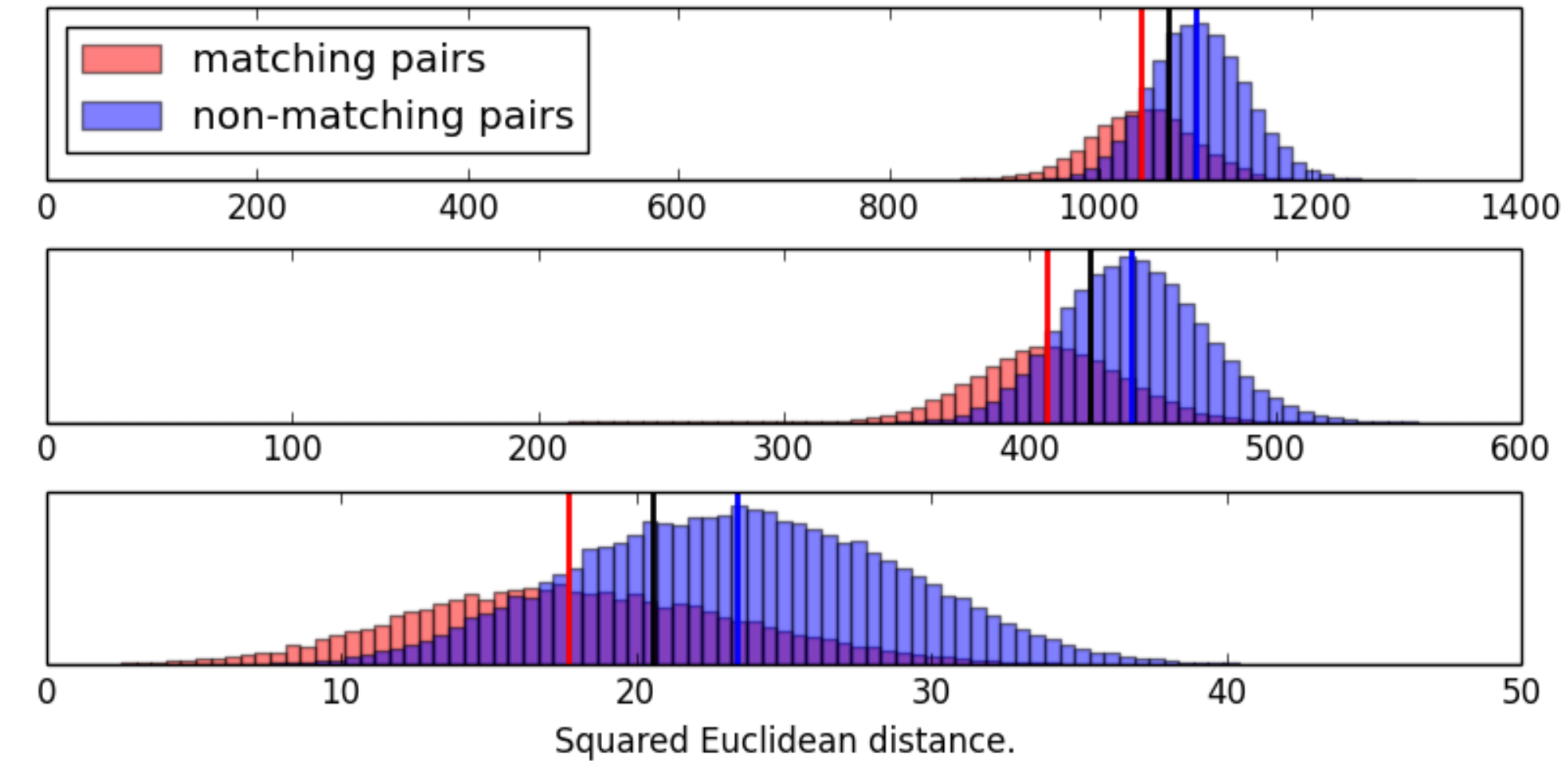}
		\caption{\footnotesize Histograms of squared Euclidean distances for 20,000 matching pairs  and corresponding 40,000 non-matching pairs for an 8192-4096(top)-2048(middle)-64(bottom) stacked RBM network. 
The red and blue vertical lines indicate the median values for the matching and non-matching pairs respectively.
The Siamese loss shared margin value $m$ is systematically set to be the mean of the two values (black vertical lines).}
	
	\label{fig:margin2}
\end{figure}

However, experiment results in Figure~\ref{fig:margin3} show that the loss function (\ref{eq:siamese1}) causes a quick drop in retrieval results.
Results with non-matching pairs alone suggest that the handling of matching pairs is responsible for the drop.
The indefinite contraction of matching pairs well beyond what is necessary to distinguish them from non-matching elements is a damaging behaviour, specially in a fine-tuning context since the network is first globally optimized with a different objective.
Figure~\ref{fig:margin2} shows that any two elements, even matching, are always far apart in high dimension.
As a solution, we propose a double-margin loss with an additional parameter affecting matching pairs:
\begin{equation}
\label{eq:siamese2}
\begin{split}
\mathcal{D}_l(\mathbf{z}_\alpha^0,\mathbf{z}_\beta^0) = &y \max(\lVert \mathbf{z}_\alpha^l - \mathbf{z}_\beta^l \rVert_2^2 - m_1, 0) \\
&+ (1-y) \max(m_2 - \lVert \mathbf{z}_\alpha^l - \mathbf{z}_\beta^l \rVert_2^2, 0)
\end{split}
\end{equation}
As shown in Figure~\ref{fig:margin1}(b), the new loss can thus be interpreted as learning ``local large-margin classifiers'' (if $m_1 \le m_2$) to distinguish between matching and non-matching elements.
In practice, we found that the two margin parameters can be set equal ($m_1 = m_2 = m$) and tuned automatically from the statistical distribution of the sampled matching and non-matching pairs (Figure~\ref{fig:margin2}).

\section{Experimental Results}
\label{sec:exp}


\subsection{Evaluation Framework}
\label{sec:eval_framework}

\paragraph{Global Descriptors.}
For the FV, we extract SIFT~\cite{Lowe04} features obtained from Difference-of-Gaussian (DoG) interest points.  
We use PCA to reduce dimensionality of the SIFT descriptor from 128 to 64 dimensions, which has shown to improve performance~\cite{Jegou_CVPR_10}.
We use a Gaussian Mixture Model (GMM) with 128 centroids, resulting in 8192 dimensions each for first and second order statistics.
Only the first-order statistics are retained in the global descriptor representation, as second-order statistics only results in a small improvement in performance~\cite{SFCV}.
The FV is $L_2$-normalized to unit-norm, after signed power normalization.
We denote this configuration as the FV feature from here-on.

DCNN features are extracted using the open-source software Caffe~\cite{Caffe} for the 7-layer AlexNet proposed for {\it ImageNet} classification in their seminal contribution~\cite{AlexNet}.
We find that layer {\it fc6} (before softmax) performs the best for image retrieval, similar to the recently reported results in~\cite{Yandex}.
We refer to this feature as the DCNN feature from here-on.

\vspace{-0.1em}
\paragraph{Training Data.}
Most schemes, including our proposed scheme, require a training step.
We use the {\it ImageNet} data set for training, which consists of 1 million images from 1000 different image categories~\cite{DengImagenet}.
We randomly sample a subset of images from {\it ImageNet}.
For the proposed deep Siamese fine-tuning scheme proposed, we use the 200K matching image pairs data set provided by {\it Yandex} in their recent work~\cite{Yandex}, consisting primarily of landmark images. 
For every matching pair, a random sample is picked to generate 2 corresponding non-matching pairs.
This training set is independent of the query and database data described next.

\begin{figure*}
	\centering
		\begin{tabular}{ @{}c@{} @{}c@{} @{}c@{} @{}c@{}}
			\includegraphics[width=2in]{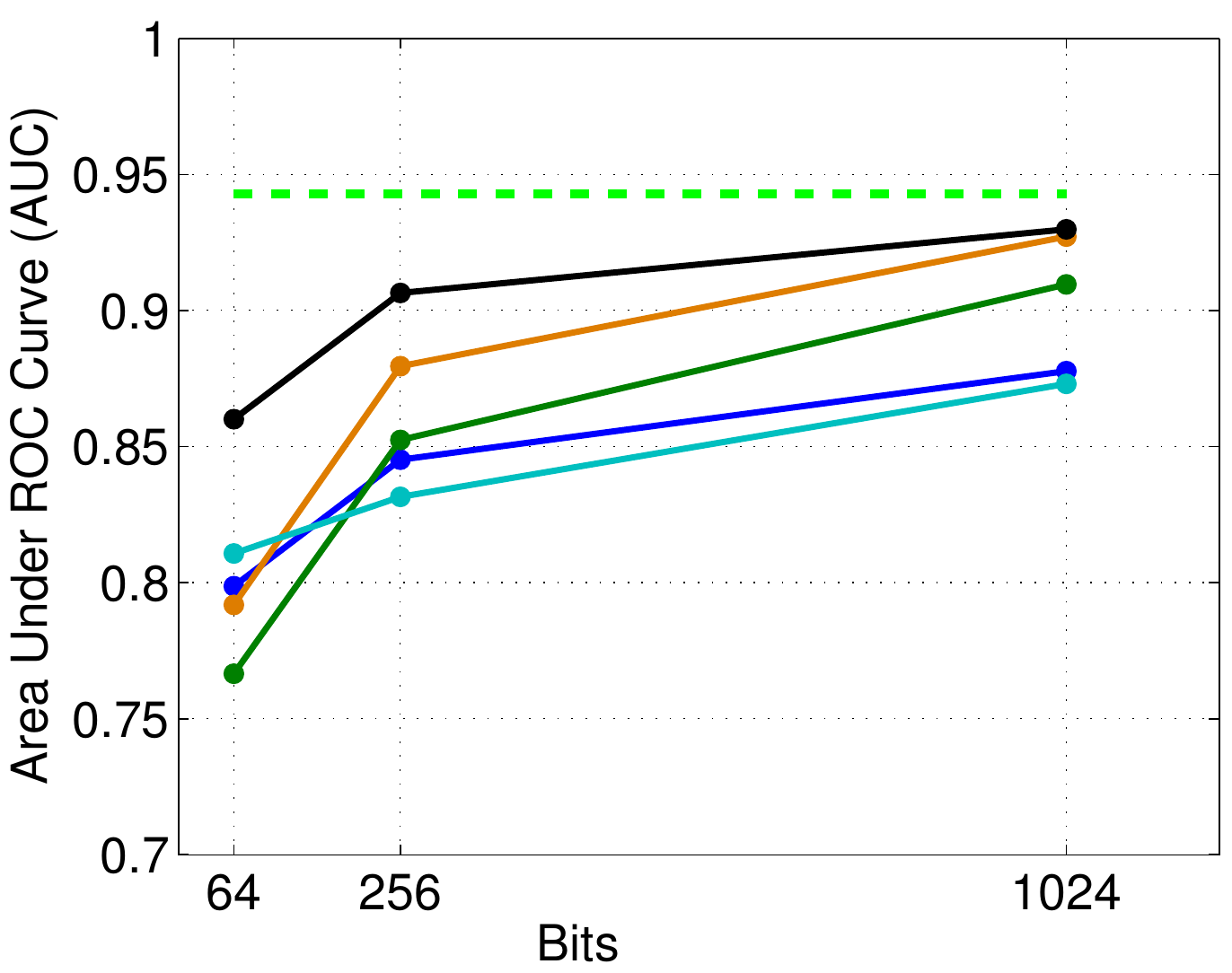} &
			\includegraphics[width=2in]{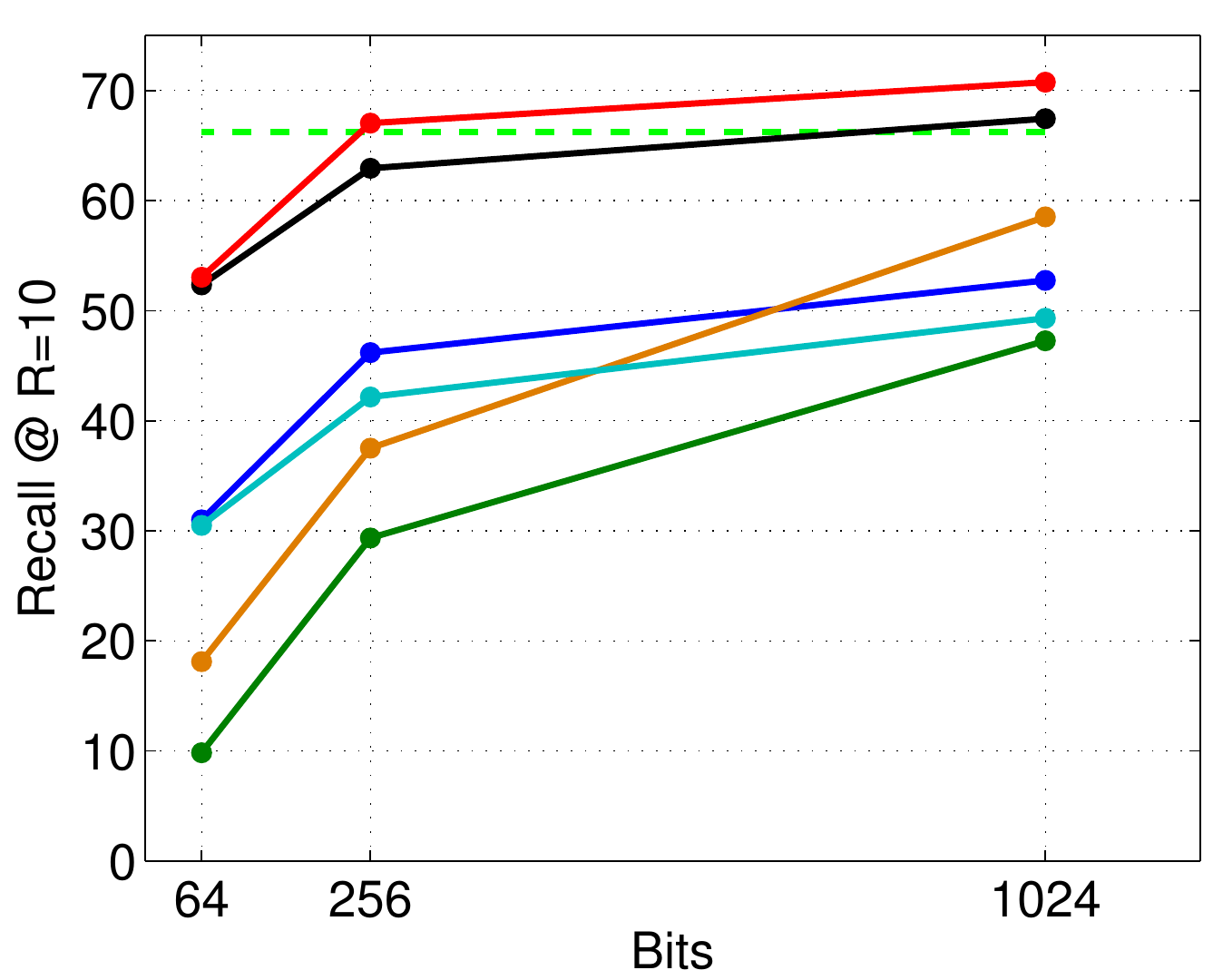} & 
			\includegraphics[width=2in]{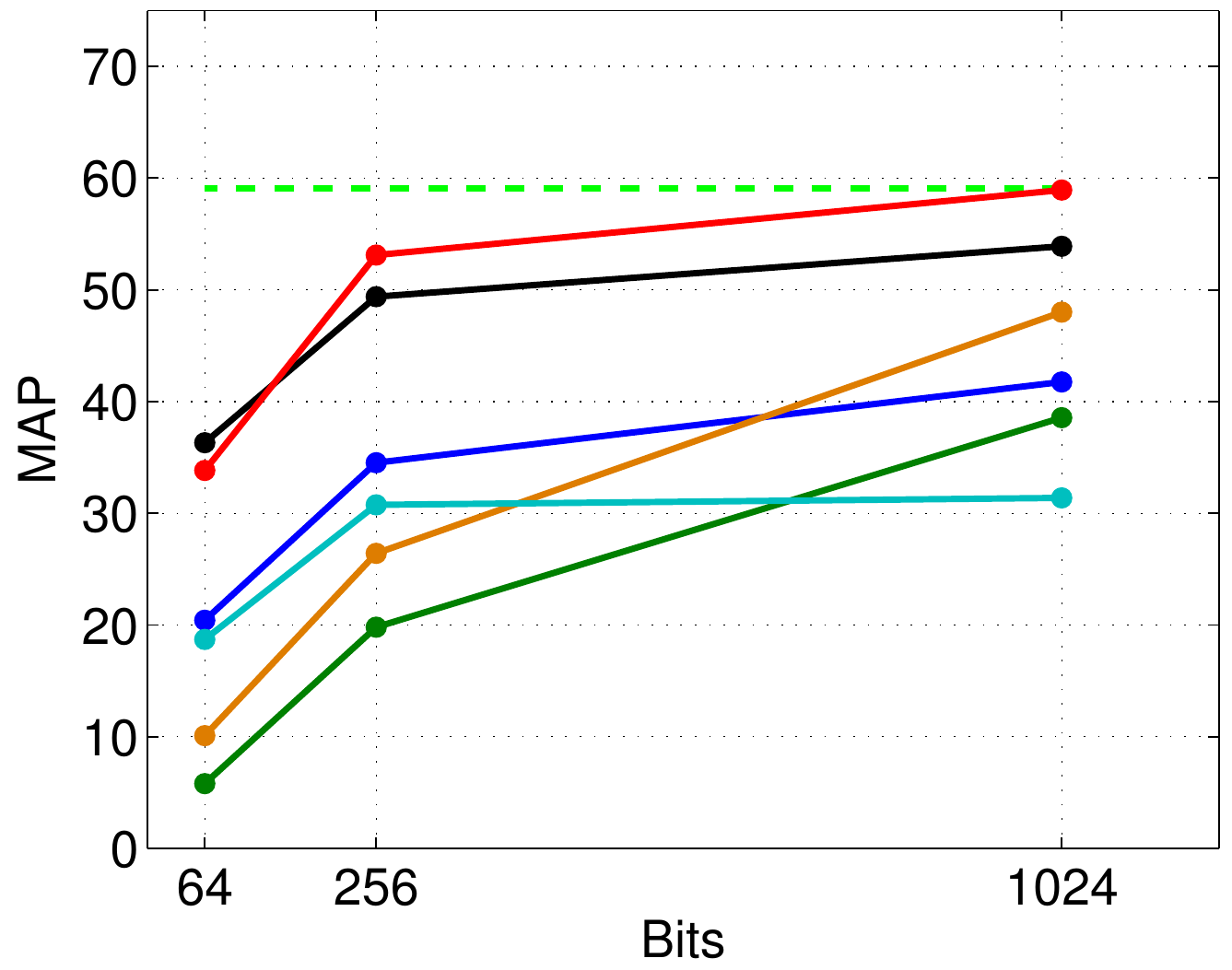} & 
			\includegraphics[width=0.8in]{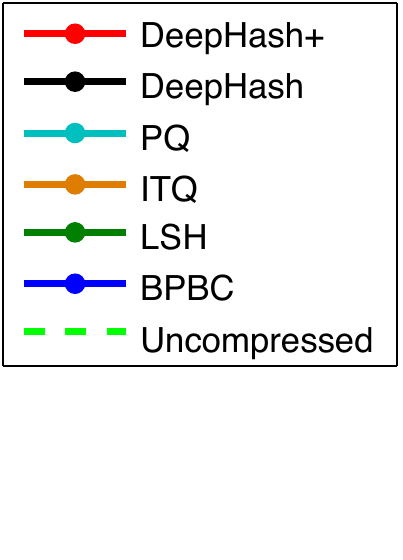}  \\ 			
			(a) & (b) & (c) &  \\
		\end{tabular}
		\caption{\footnotesize 
		Comparing AUC, Recall and MAP performance of different schemes at varying $b$ in (a),(b) and (c) respectively. {\it Holidays} and FV are used for retrieval experiments, and {\it SMVS} for AUC. DeepHash outperforms all schemes.  Also, the performance ordering of schemes is largely consistent between AUC results and retrieval results, both MAP and Recall. AUC can be used for fast optimization of parameters.
		}	
		\label{fig:auc_all}
		
\end{figure*}

\vspace{-0.1em}
\paragraph{Testing Data.}
We use 4 popular data sets for small scale experiments: {\it Oxford} (55 queries, 5062 database images)~\cite{Philbin07},  {\it INRIA Holidays} (500 queries, 991 database images)~\cite{Jegou08}, Stanford Mobile Visual Search {\it Graphics}  (1500 queries, 1000 database images)~\cite{SVMSDataSet,MPEGDataset2} and University of Kentucky Benchmark (UKB) (10200 queries, 10200 database images)~\cite{Nister06}.
For large-scale retrieval experiments, we present results on {\it Holidays} and {\it UKB} data sets, combined with the 1 million MIR-FLICKR distractor data set~\cite{mirflickr}.

\begin{figure}
	\centering
		\begin{tabular}{ @{}c@{} @{}c@{} }
			\includegraphics[width=1.7in]{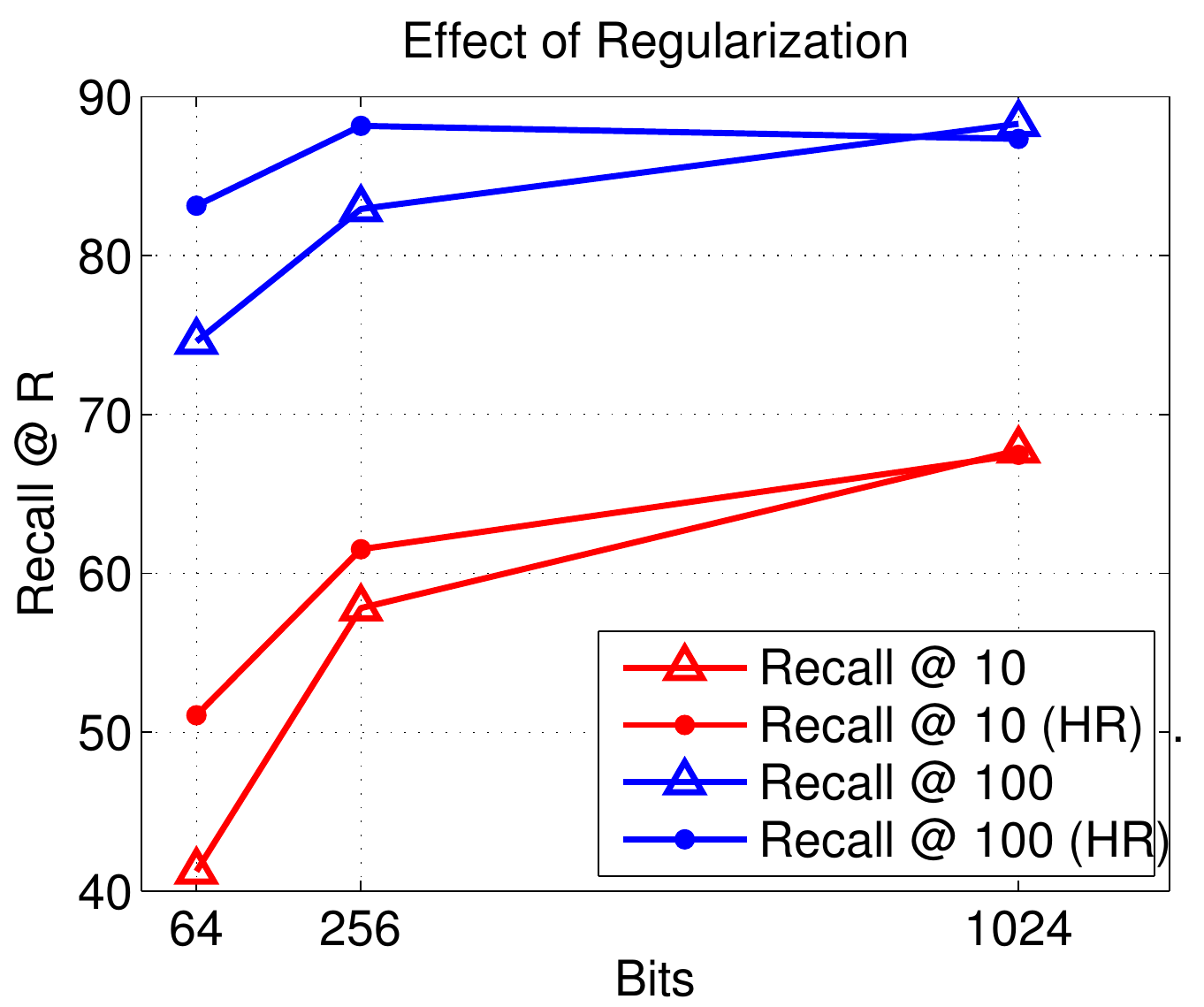} &
			\includegraphics[width=1.7in]{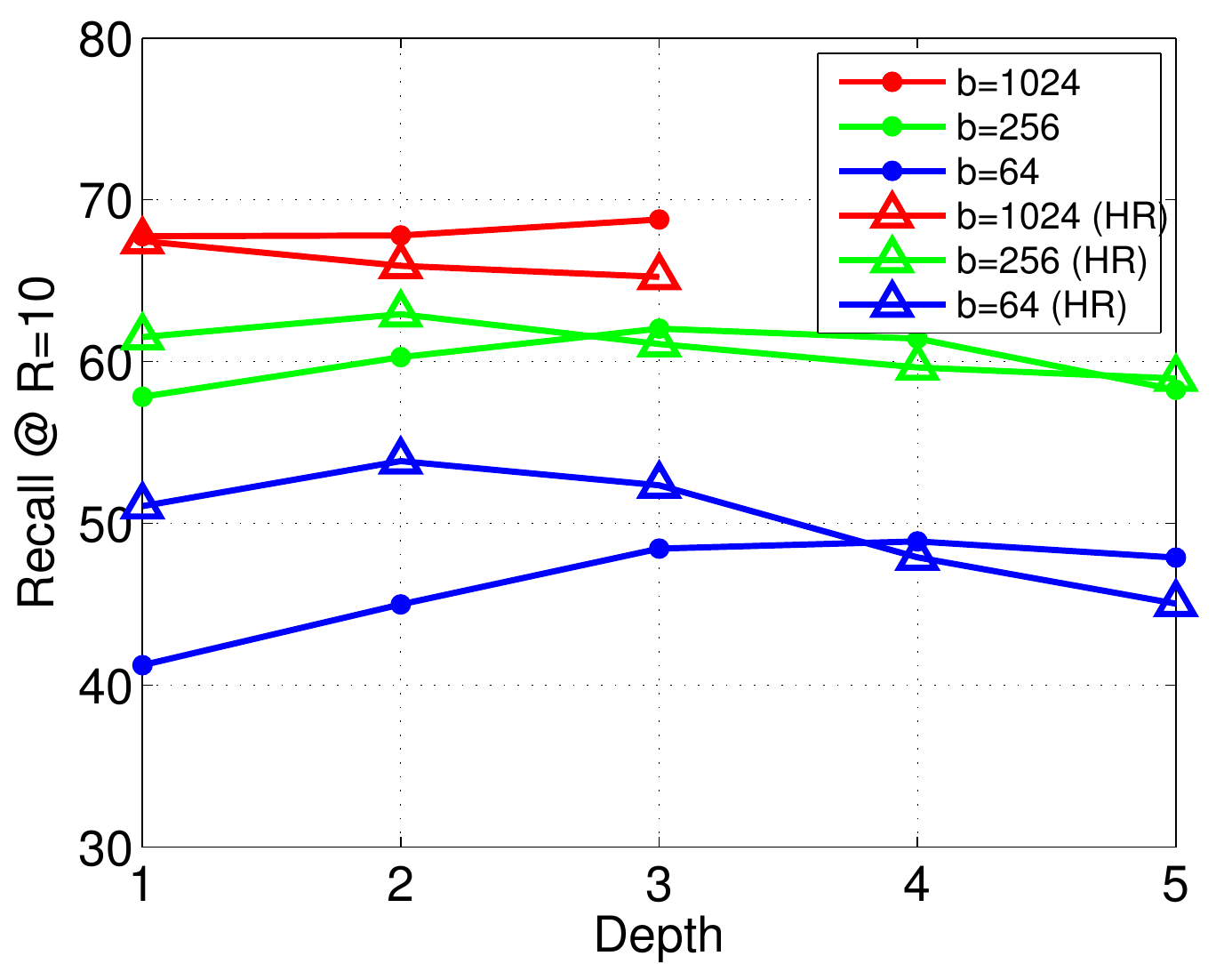} \\
			(a) & (b) \\
		\end{tabular}
		\caption{\footnotesize Hashing FV for {\it Holidays}. (HR) refers to schemes trained with hashing regularization. (a) Hashing regularization improves performance significantly for single layer models 8192-$b$ as $b$ is decreased. (b) Recall improves as depth is increased for lower rate points $b=64$ and $b=256$. With regularization, we can achieve the same or better recall at lower depth. 
		}	
		\label{fig:depth_sparsity}
		
\end{figure}

\vspace{-0.1em}
\paragraph{Comparisons.}
We compare several state-of-the-art schemes.
Some have been proposed for lower dimensional vectors like SIFT and GIST, but we evaluate their performance on both FV and DCNN features.

\squishlist

\item {\it ITQ}~\cite{ITQ}. 
For the Iterative Quantization (ITQ) scheme, the authors propose signed binarization after applying two transforms: first the PCA matrix, followed by a rotation matrix, which minimizes the quantization error of mapping PCA-transformed data to the vertices of a zero-centered binary hypercube.

\item {\it BPBC}~\cite{BPBC}.
Instead of the large PCA projection matrices used in~\cite{ITQ}, the authors apply bilinear projections, which require far less memory. 

\item {\it LSH}~\cite{RandomProjections}.
LSH is based on random unit-norm projections followed by signed binarization.

\item {\it PQ}~\cite{PQFisher}.
For FV and Product Quantization, we consider blocks of dimensions $D=64, 256$ and $1024$, and train $K=256$ centroids for each block, resulting in $b=64, 256$ and $1024$ bit descriptors respectively.
For DCNN, we consider blocks of dimensions $D=32, 128$ and $1024$, with $K=256$ centroids, resulting in the same bitrates.
Here, we do not apply Random Rotations, or PCA before applying PQ~\cite{PQFisher}.
Such preprocessing can be applied to other schemes too.
This is not a binary hashing scheme and only included for reference.

\squishend

We ignore Spectral Hashing~\cite{SpectralHashing} due to its inferior performance on FV in~\cite{Perronnin_CVPR_10}.

\subsection{DeepHash Experiments}
\label{sec:rbm_experiments}

\paragraph{Hashing Regularization.}
In Figure~\ref{fig:depth_sparsity}(a), we show the effect of applying regularization proposed in Section~\ref{sec:srbm} on a single layer RBM 8192-$b$, for $b=64,256,1024$.
The {\it Holidays} data set and FV features are chosen.
Hashing regularization improves performance significantly, $\sim$10$\%$ absolute recall @ $R=10$ at low-rate point $b=64$.
The performance gap increases as rate decreases.
This is intuitive as the regularization pushes the network towards keeping half the bits alive and equiprobable (across hashes), with its effect being more pronounced at lower rates.

\vspace{-0.1em}
\paragraph{Depth.}
In Figure~\ref{fig:depth_sparsity}(b), we plot recall @ $R=10$ for the {\it Holidays} data set and {\it FV} features, as depth is increased for a given rate point $b$.
For $b=1024$, we consider configurations 8192-1024, 8192-4096-1024, and 8192-4096-2048-1024 corresponding to depth $1,2,3$ respectively. 
For rate points $b=64$ and $256$, similar configurations of varying depth are chosen.
We observe that, with no regularization, recall improves as depth is increased for $b=256$ and $b=64$, with optimal depth of 3 and 4 respectively, beyond which performance drops.
At higher rates of $b=1024$ and beyond, increasing depth does not improve as performance saturates.
For hashing, a sweet spot in performance for the depth parameter is observed for each rate point, as deeper networks can cause performance to drop due to loss of information over the layers.
Similar trends are obtained for recall @ $R=100$.
Importantly, we observe that with the proposed regularization, we can achieve the same performance with lower depth at each rate point.
This is critical, as lower the depth, the faster the hash generation, and lower the memory requirements.

\vspace{-0.1em}
\paragraph{Fine-Tuning.}

\begin{table}[h]
\centering 
\caption{\footnotesize Retrieval results before and after Siamese fine-tuning, with corresponding differences.
The stacked RBM network (8192-4096-2048-64) is trained with Fisher descriptors from the ImageNet data set.  
Fine-tuning consistently improves retrieval results at any bit-rate.
}
\footnotesize
\scalebox{1.}{
\begin{tabular}{|c|c|c|c|c|c|c|c|}
	\hline
	\multirow{2}{*}{} & \multirow{2}{*}{Layer} & \multicolumn{3}{ c| }{Recall @ R=10} & \multicolumn{3}{ c| }{Recall @ R=100}\\
	\cline{3-8}
	& & bef. & aft. & diff. & bef. & aft. & diff.\\
	\hline
	\multirow{3}{*}{\rotatebox[origin=c]{90}{Holidays}} & 4096 & 70.83 & 73.67 & \bf{2.84} & 89.92 & 91.40 &\bf{1.48}\\
	\cline{2-8}
	& 2048 & 67.74 & 71.12 & \bf{3.38} & 88.77 & 92.04 &\bf{3.27}\\
	\cline{2-8}
	& 64 & 52.06 & 53.04 & \bf{0.98} & 80.38 & 83.91 &\bf{3.53}\\
	\hline
	\multirow{3}{*}{\rotatebox[origin=c]{90}{Oxford}} & 4096 & 19.38 & 21.73 & \bf{2.35} & 41.09 & 45.19 &\bf{4.10}\\
	\cline{2-8}
	& 2048 & 14.32 & 17.23 & \bf{2.91} & 36.03 & 41.03 &\bf{5.00}\\
	\cline{2-8}
	& 64 & 10.69 & 12.01 & \bf{1.32} & 23.75 & 29.99 &\bf{6.24}\\
	\hline
	\multirow{3}{*}{\rotatebox[origin=c]{90}{UKB}} & 4096 & 79.22 & 82.22 & \bf{3.00} & 92.04 & 93.73 &\bf{1.69}\\
	\cline{2-8}
	& 2048 & 75.62 & 79.37 & \bf{3.75} & 90.79 & 92.82 &\bf{2.03}\\
	\cline{2-8}
	& 64 & 47.94 & 49.25 & \bf{1.31} & 73.02 & 73.94 &\bf{0.92}\\
	\hline
\end{tabular}
}
\label{table:finetuningres}
\end{table}

Table~\ref{table:finetuningres} provides detailed retrieval results for a 3-layer model before and after Siamese fine-tuning.
The results show consistent improvements with every training data set and at any bit-rate with a global average difference of 2.78\% (up to 6.24\%).  
The difference is more significant at higher recall rates with an average of 2.43\% @ R=10 compared to 3.13\% @ R=100.
They are however quite comparable when relative improvement rate is considered: 7.46\% @ R=10 and 7.24\% @ R=100 relatively. 

We notice differences across test sets with improvements on the {\it Oxford} set being more pronounced.
The {\it Yandex} data set used for fine-tuning is made with matching pairs of landmark structures which can explain the better performance of the {\it Oxford} data set made of buildings only.
The systematic improvements on all sets are nevertheless evidence of the high transferability of both unsupervised training and semi-supervised fine-tuning.

\vspace{-0.1em}
\paragraph{Fast Optimization with ROC Experiments.}

The Stanford Mobile Visual Search ({\it SMVS}) data set~\cite{SVMSDataSet} contains a list of 16,319 matching image pairs, comprising a wide range of object categories.
We extract FV for matching and non-matching pairs from the {\it SMVS} data set, hash the data to different rate points, and compute the Receiver Operating Characteristic (ROC) curve.
In Figure~\ref{fig:auc_all}(a), we plot ROC Area Under Curve (AUC) for different schemes for the {\it SMVS} data set. 
In Figure~\ref{fig:auc_all}(b) and ~\ref{fig:auc_all}(c), we plot recall @ 10 and MAP on the {\it Holidays} data set. 
The retrieval performance of a scheme at a given database size depends on the ROC curve at different TPR/FPR operating points, as shown in~\cite{DavidChenThesis}.
Low FPR points are important~\cite{DavidChenThesis,REVV1}.
We observe that the Area Under Curve (AUC) results predict well the performance ordering (MAP and Recall) of different schemes for retrieval experiments.
The retrieval and AUC experiments are performed on very different data sets, but the AUC results generalize well, and are used for fast optimization of parameters.

\vspace{-0.1em}
\paragraph{DeepHash Parameters.}

For RBM learning, we set the learning rate to 0.001 for the weight and bias parameters, momentum to 0.9, and ran the training on 150,000 images from the {\it ImageNet} data set for a maximum 30 epochs. 
Binary descriptors for the first layer are generated by subtracting the mean for each data set.
For each rate point, we consider a set of models with dimensionality progressively reduced by a factor of 2 from the starting representation for FV and DCNN respectively.
The best model is chosen based on greedy optimization of AUC on the {\it SMVS} data set, which works well as seen in detailed experimental results of Section~\ref{sec:retrieval_results}.
The chosen architectures are described in Table~\ref{tab:rbm_params}.
The depth of the network increases as hash size decreases.
Each target setting requires several hours to train on a modern CPU.

\begin{table}
\centering
\begin{tabular}{|c|c|c|}
\hline
Bits 		& 	FV  				& 	DCNN  			\\
\hline
1024 		& 	8192-1024 			& 	4096-1024 		\\
\hline
256 		& 	8192-4096-256 		& 	4096-2048-256 		\\
\hline
64 			& 	8192-4096-2048-64 	& 	4096-2048-1024-64 		\\
\hline
\end{tabular}
\caption{DeepHash Architecture}
\label{tab:rbm_params}
\end{table}

\subsection{Retrieval Experiments}
\label{sec:retrieval_results}

\begin{figure}
	\centering 
		\includegraphics[width=2.5in]{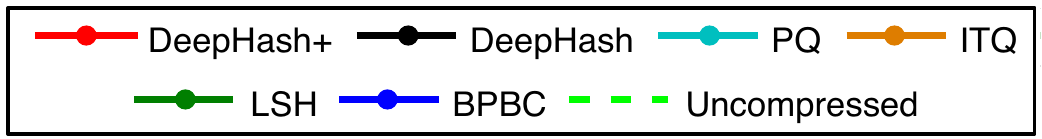} 
		\begin{tabular}{ @{}c@{} @{}c@{} }
					
			\includegraphics[width=1.7in]{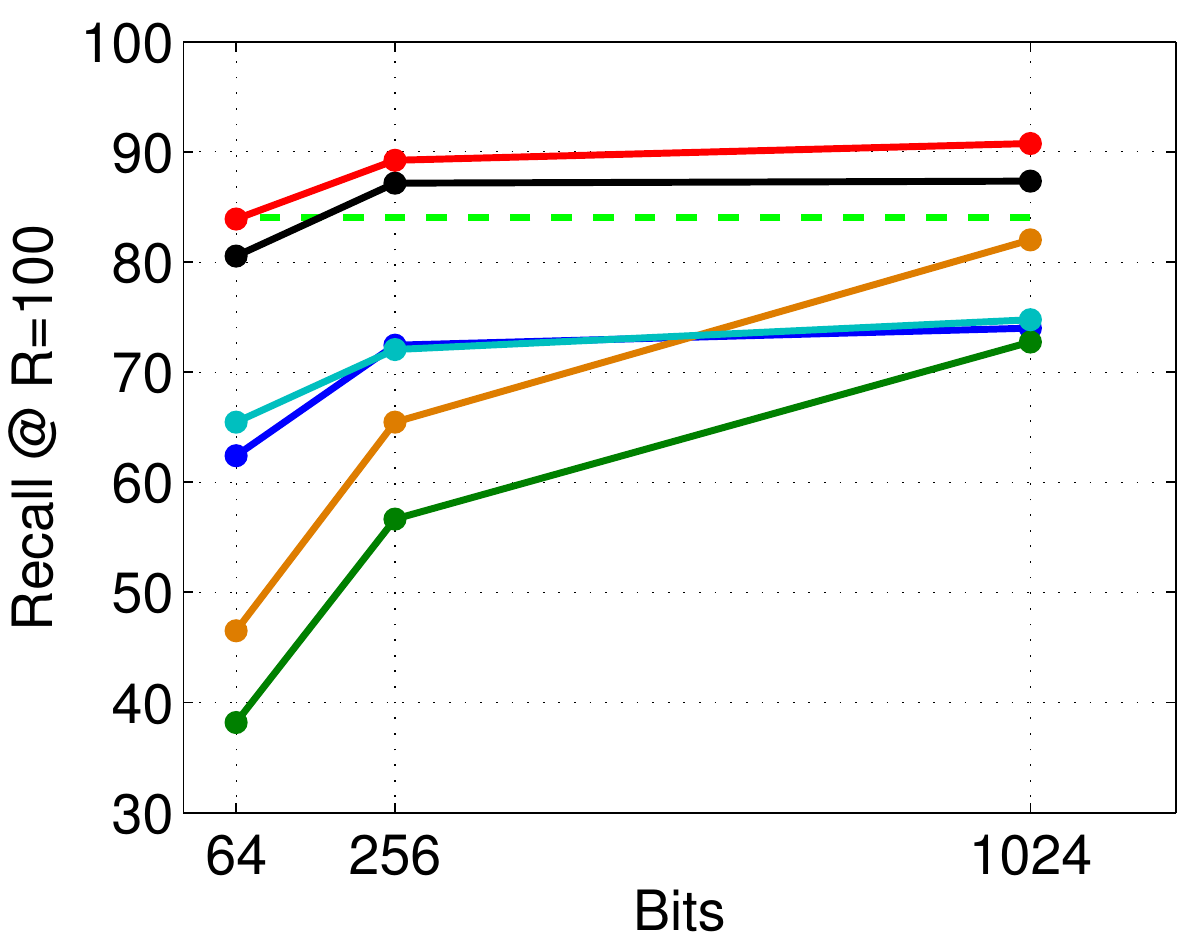} &
			\includegraphics[width=1.7in]{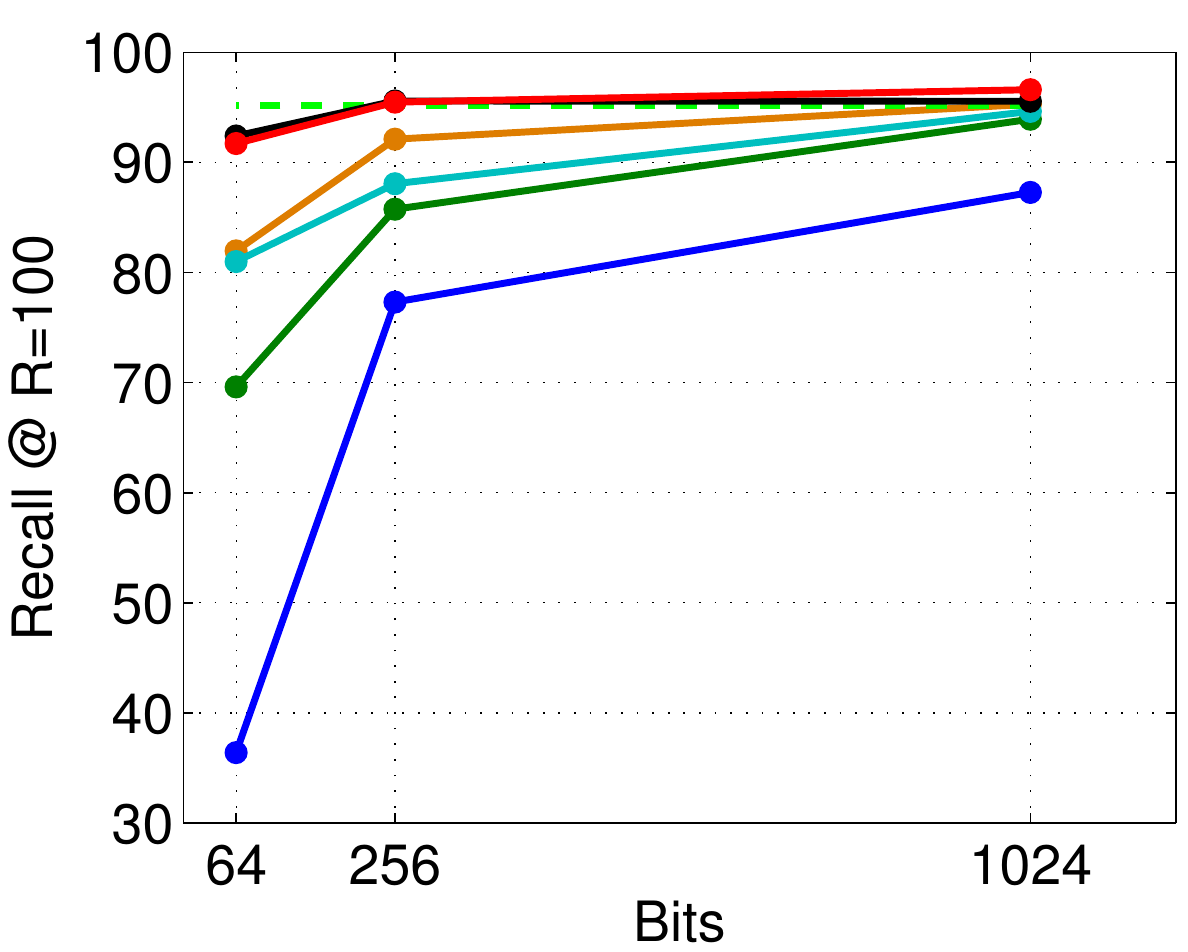} \\ 			
			{\it (a) FV, Holidays} & {\it (b) DCNN, Holidays} \\

			\includegraphics[width=1.7in]{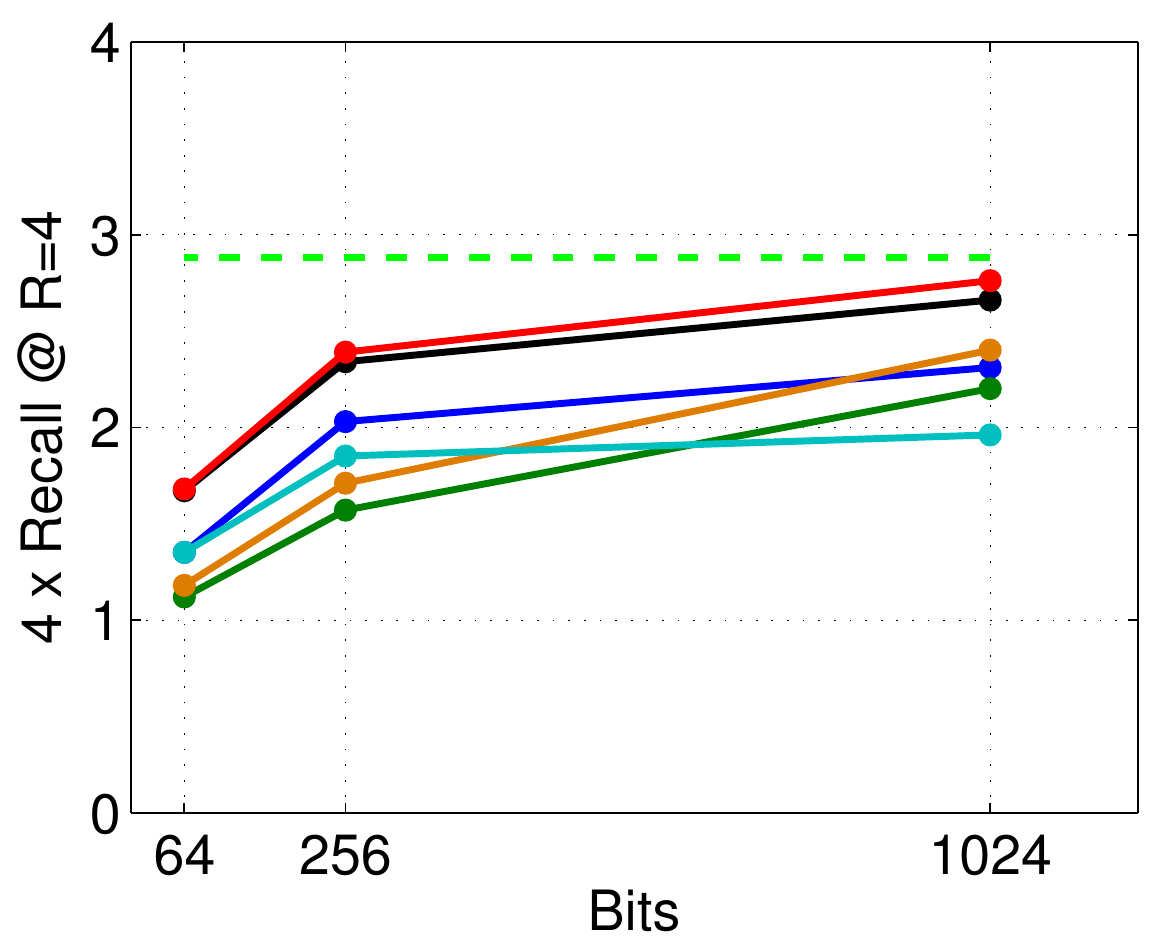} &
			\includegraphics[width=1.7in]{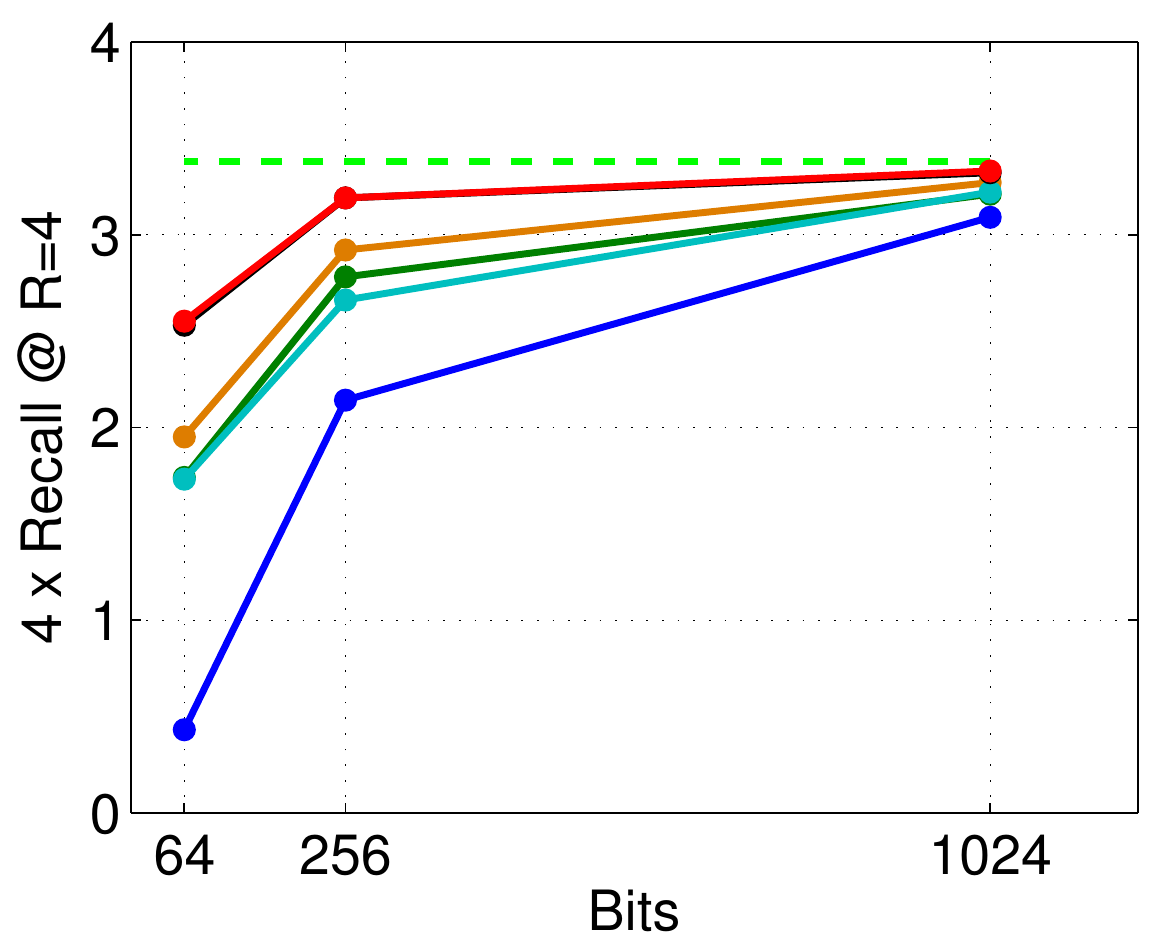} \\ 
			{\it (c) FV, UKB} & {\it (d) DCNN, UKB} \\

			\includegraphics[width=1.7in]{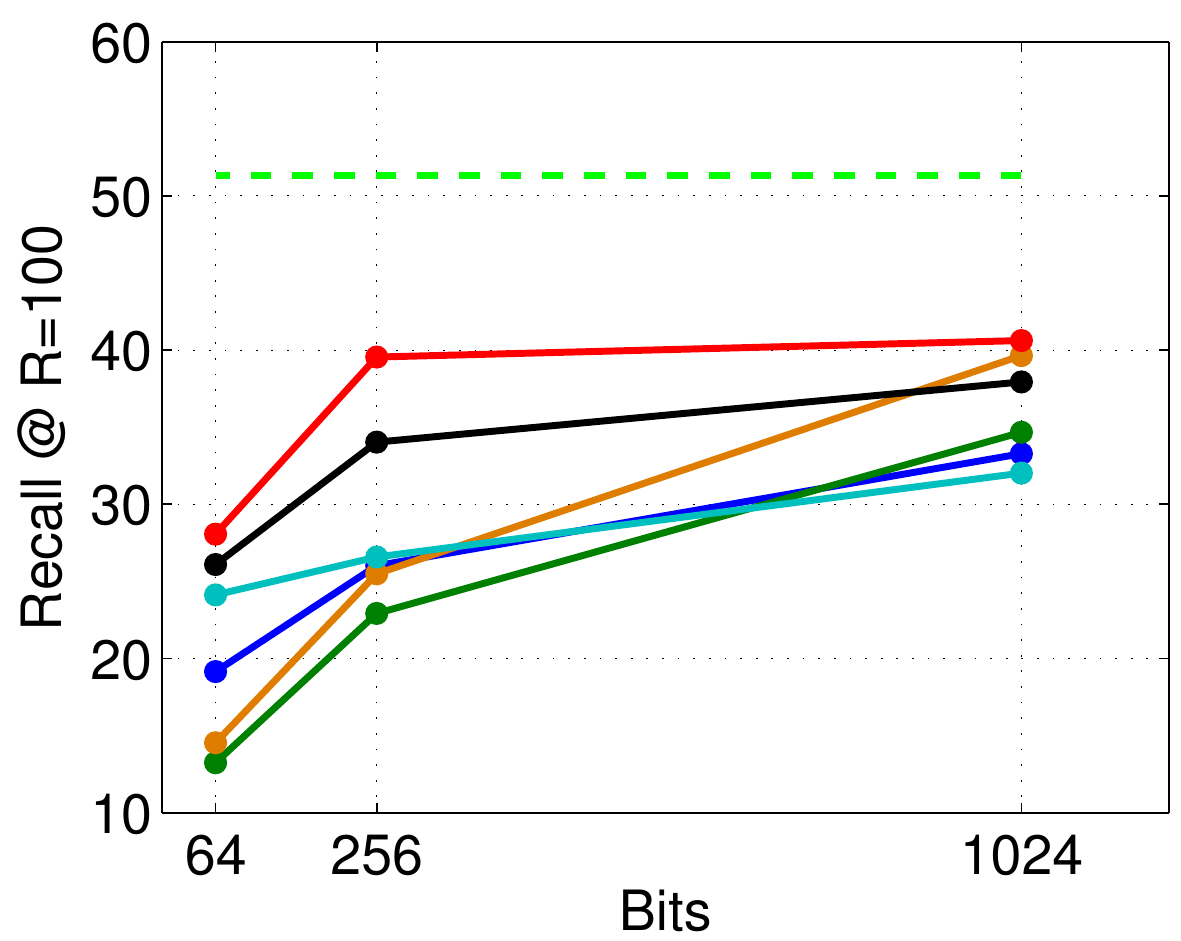} &
			\includegraphics[width=1.7in]{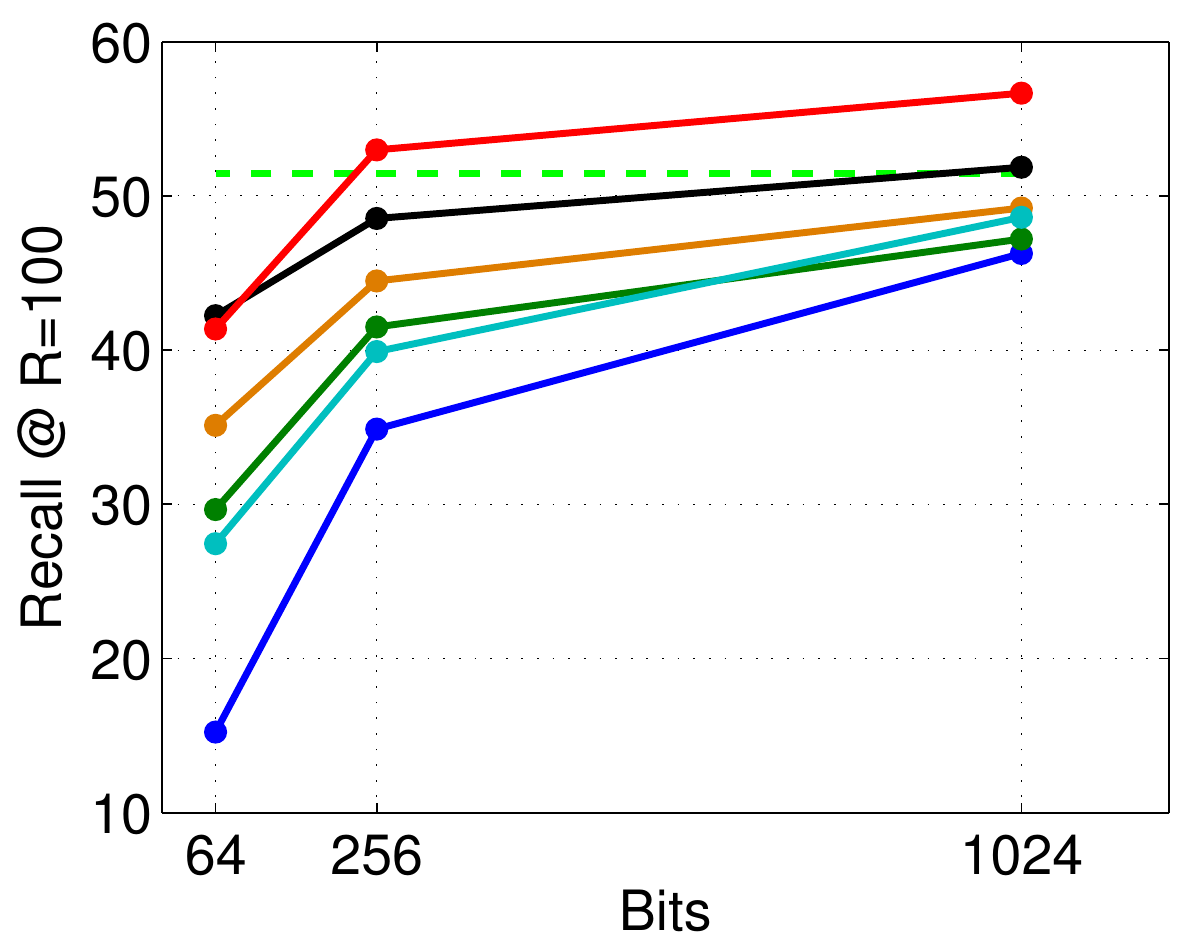} \\ 
			{\it (e) FV, Oxford} & {\it (f) DCNN, Oxford} \\
			
			\includegraphics[width=1.7in]{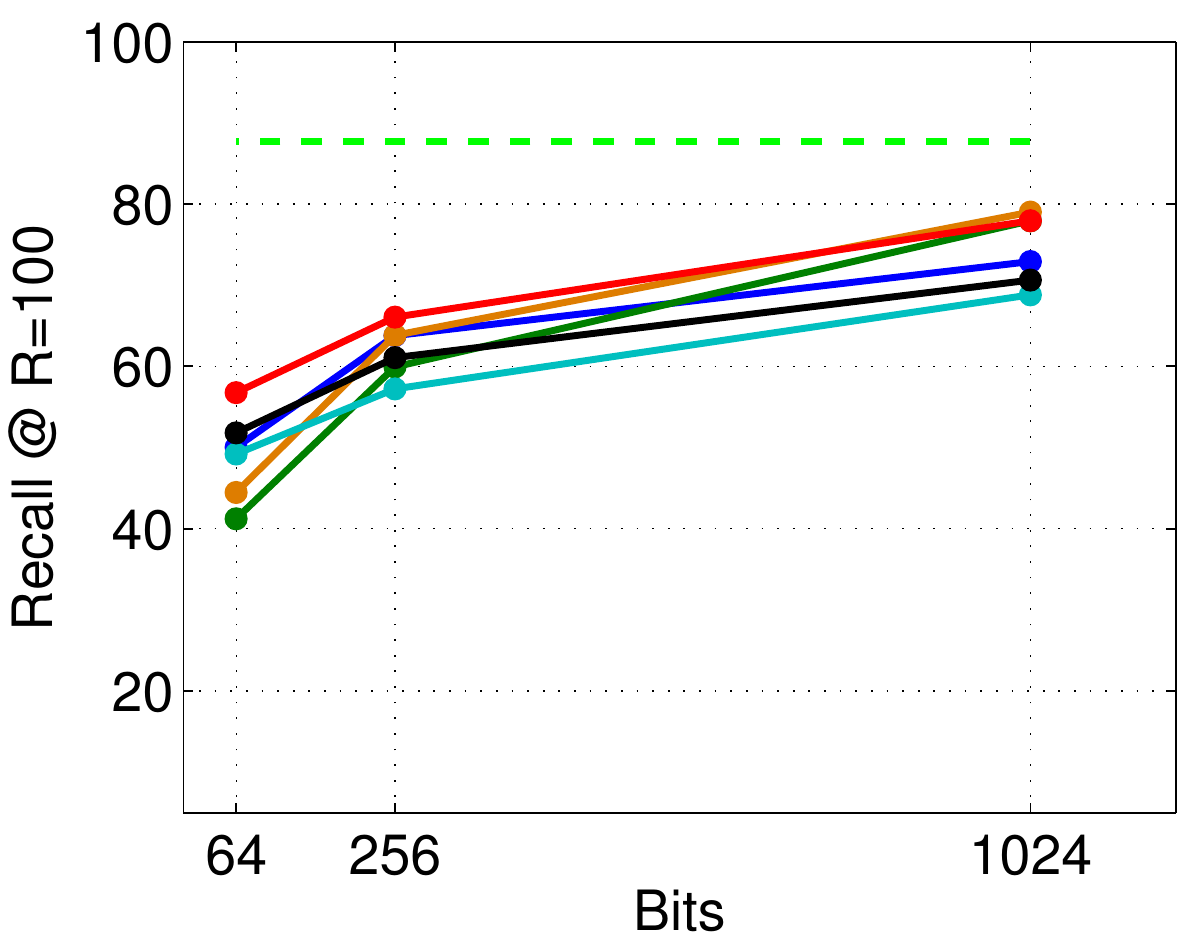} &
			\includegraphics[width=1.7in]{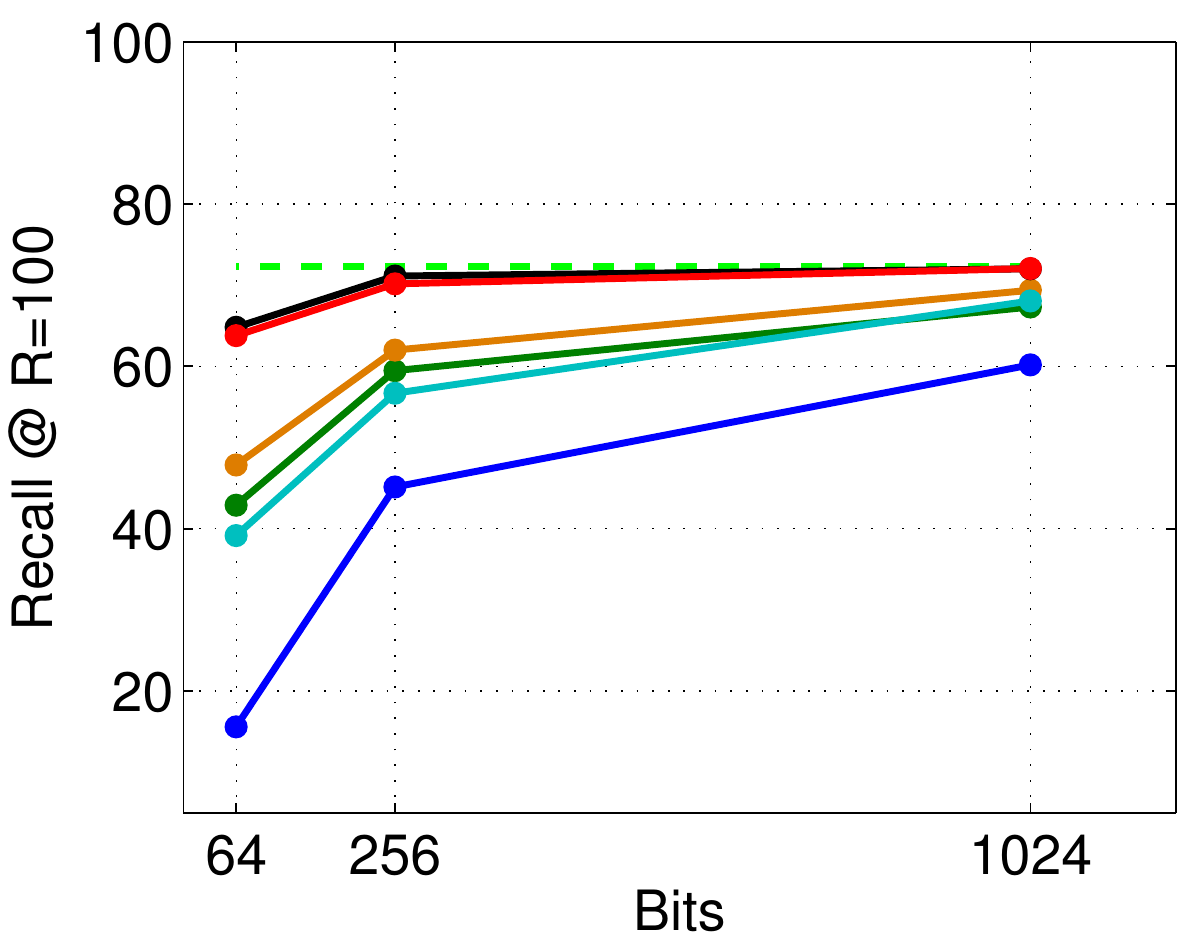} \\ 
			{\it (g) FV, Graphics} & {\it (h) DCNN, Graphics} \\

		\end{tabular}
		\caption{\footnotesize Small-scale retrieval results. {\it DeepHash} outperforms other schemes by a significant margin.
		}	
		\label{fig:retrieval_small_scale}
\end{figure}

\begin{figure}
	\centering 
		\includegraphics[width=2.5in]{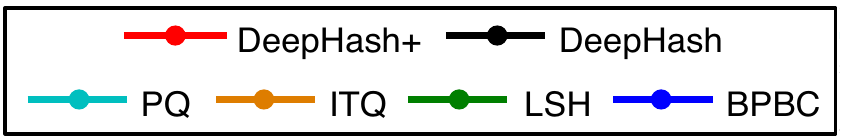} 

		\begin{tabular}{ @{}c@{} @{}c@{} }
			\includegraphics[width=1.7in]{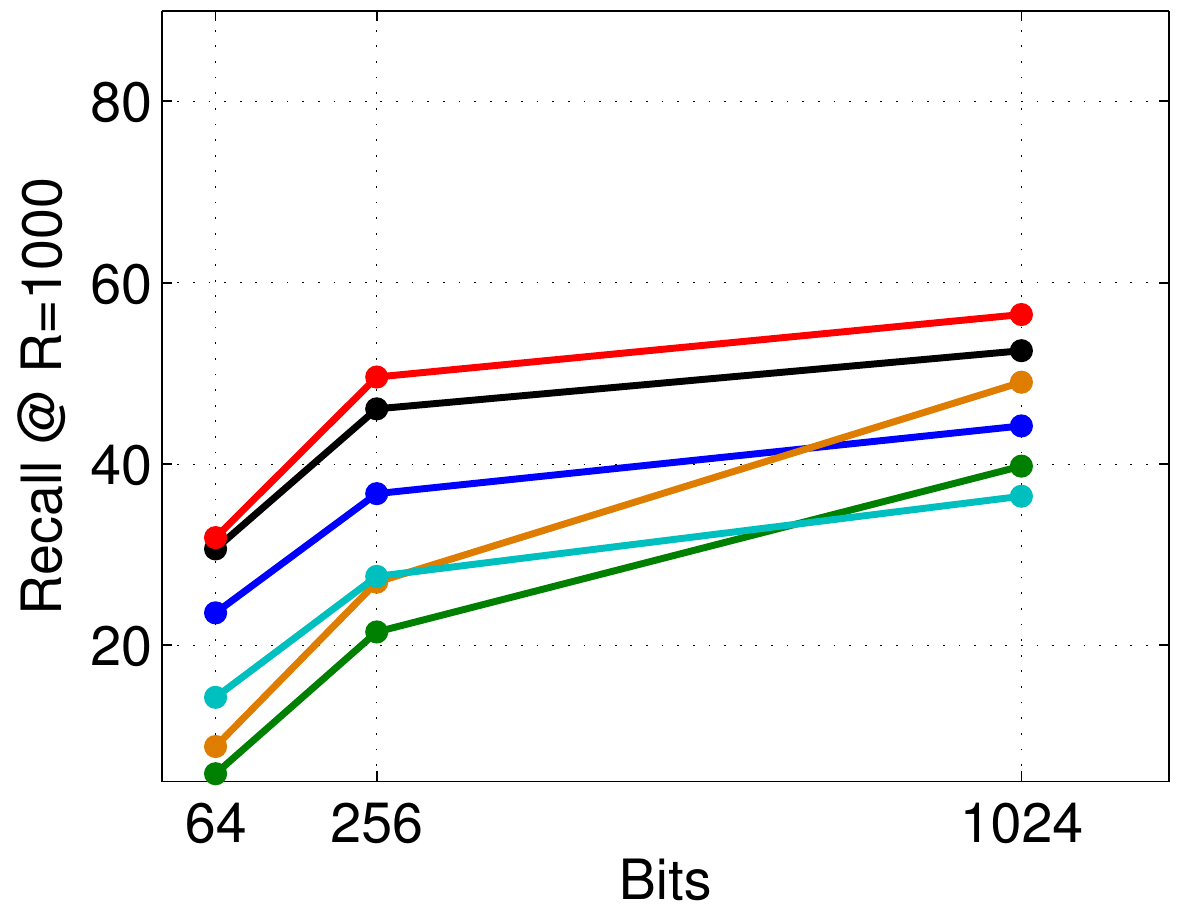} &
			\includegraphics[width=1.7in]{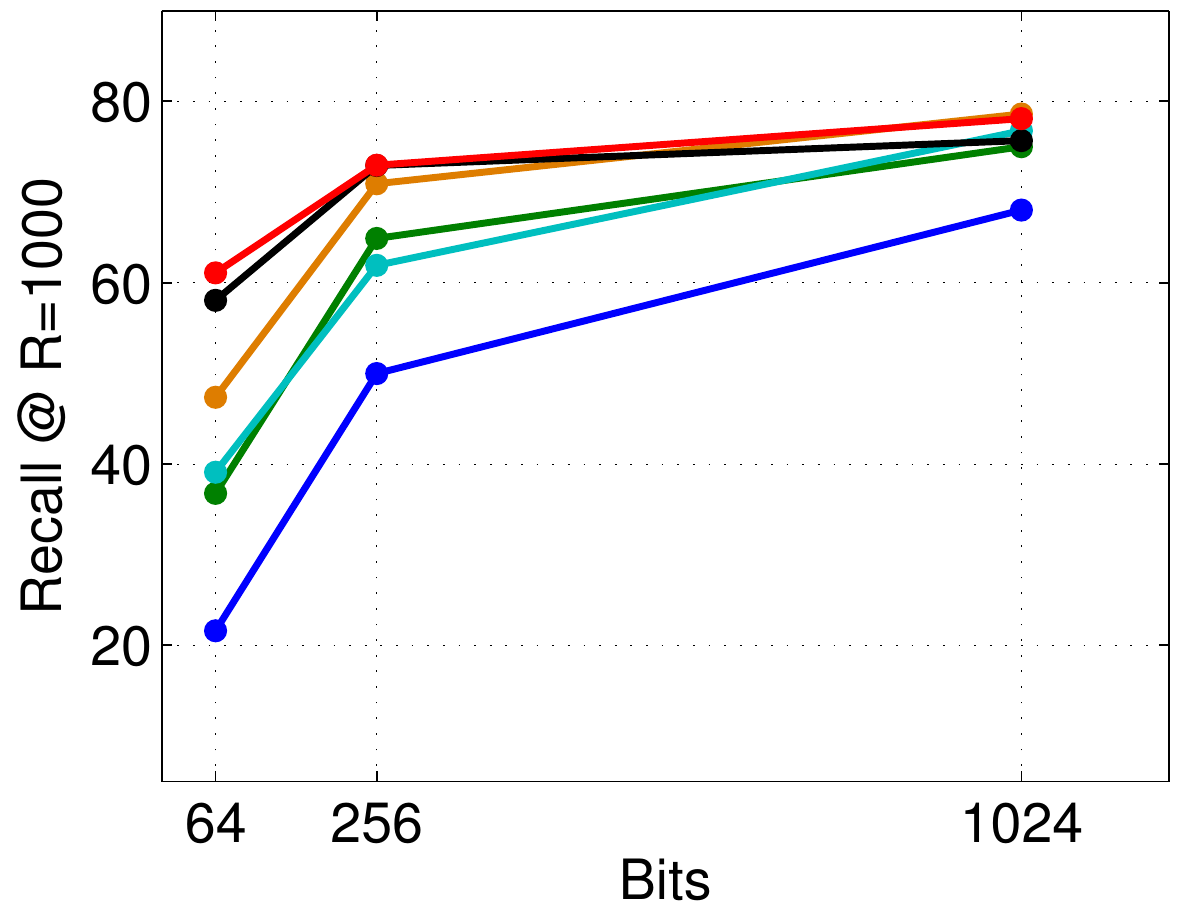} \\
			
			{\it FV, Holidays+1M} & {\it DCNN, Holidays+1M} \\
			
			\includegraphics[width=1.7in]{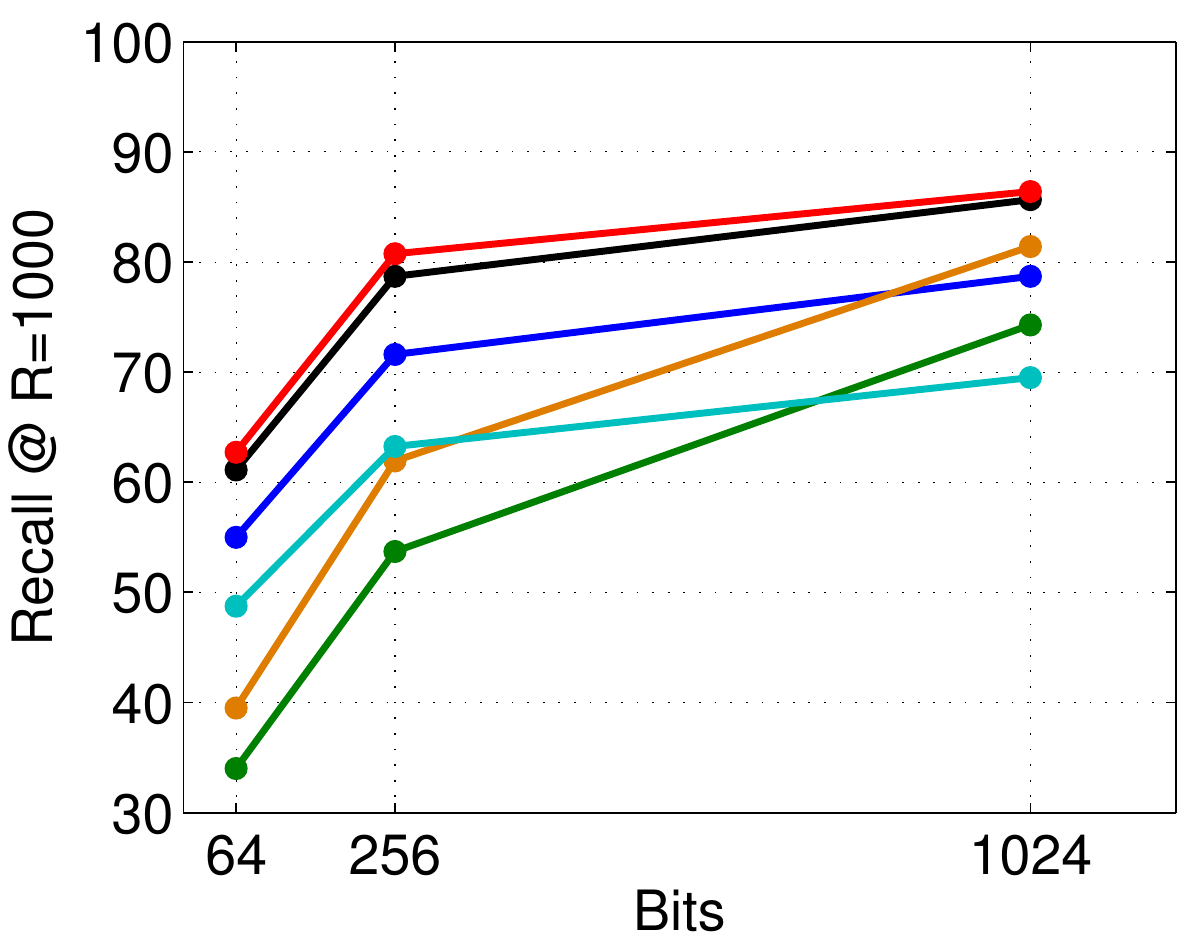} & 
			\includegraphics[width=1.7in]{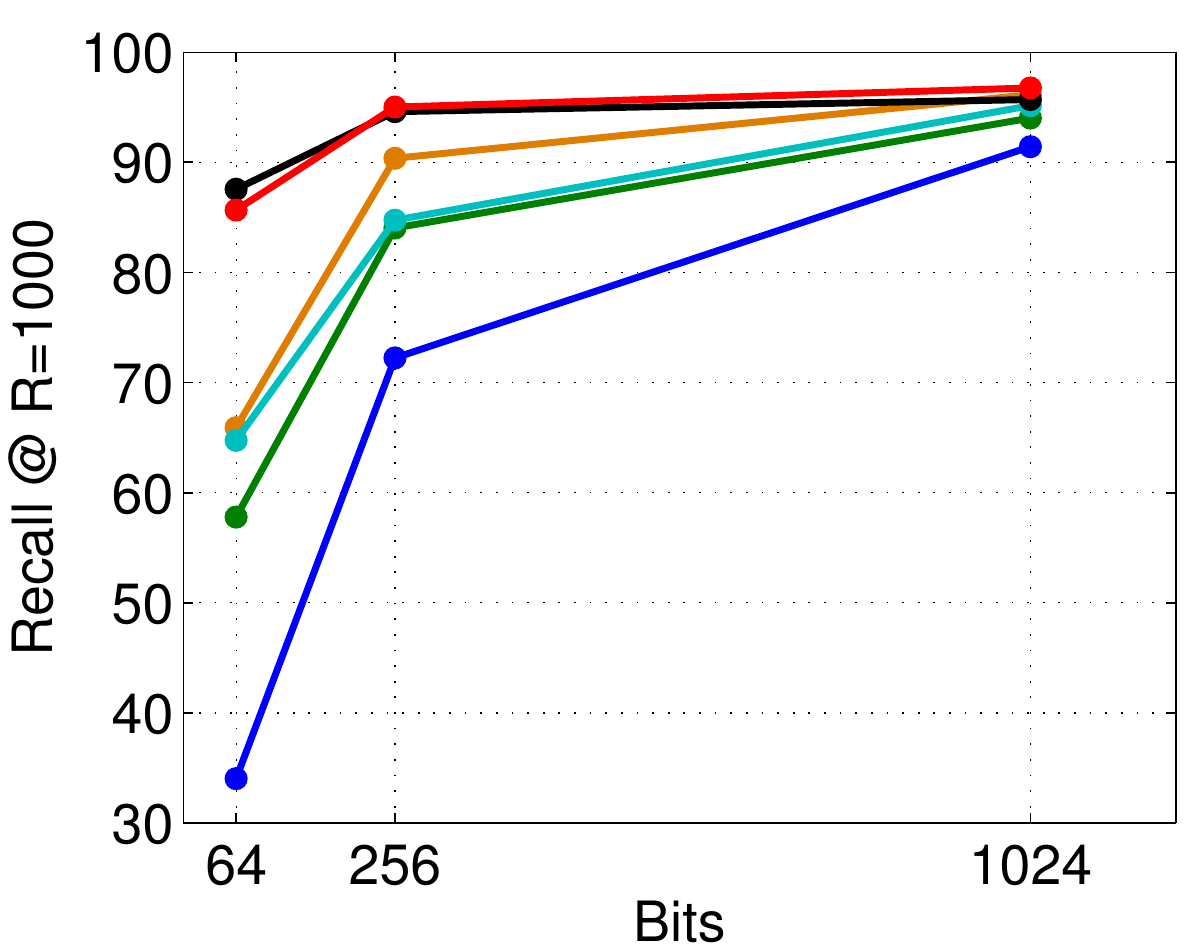} \\
			
			{\it FV, UKB+1M} & {\it DCNN, UKB+1M} \\
			
		\end{tabular}
		\caption{\footnotesize 
		Large-scale retrieval results (with 1M distractor images) for different compression schemes. {\it DeepHash} outperforms other schemes at most rate points and data sets.
		}	
		\label{fig:retrieval_large_scale}
\end{figure}

We present retrieval results using FV and DCNN features in Figure~\ref{fig:retrieval_small_scale} and Figure~\ref{fig:retrieval_large_scale}.
For instance retrieval, it is important for the relevant image to be present in the first step of the pipeline, matching global descriptors, so that a Geometric Consistency Check (GCC)~\cite{Fischler81} step can find it subsequently. 
We present recall @ typical operating points, R = 100 and R = 1000 for small and large data sets respectively. 
For {\it UKB} small experiments, we plot 4$\times$ recall @ $R=4$ to be consistent with the literature.
We refer to hashes before and after fine-tuning as DeepHash and DeepHash+ respectively in all figures.
We refer to deep hashes based on DCNN and FV features as DCNN-DeepHash and FV-DeepHash respectively.

\vspace{-0.1em}
\paragraph{Performance of DeepHash.}
For DCNN and FV features, the proposed DeepHash outperforms state-of-the-art schemes by a significant margin on all data sets.
The statistics of FV and DCNN features are very different. 
FV are dense descriptors with zero blocks corresponding to centroids not visited, while deep DCNN features tend to be sparse.
Our method works well for both types of features.

For the retrieval experiments in Figure~\ref{fig:retrieval_small_scale}, there is up to 20$\%$ improvement in absolute recall at $b=64$ bits compared to the second performing scheme.
Up to 15$\%$ improvement is seen at $b=256$, which can be a practical rate point for many applications, as there is only a marginal drop in performance for DCNN features compared to uncompressed features.
Similar trends are obtained for recall @ $R=10$ and MAP, as seen by comparing {\it Holidays} results in Figure~\ref{fig:auc_all}(b),(c) and Figure~\ref{fig:retrieval_small_scale}(a), with a higher gap for larger $R$.
Consistent trends are also obtained for the large-scale retrieval results in Figure~\ref{fig:retrieval_large_scale}.

The performance ordering of other schemes depends on the bitrate and type of feature, while DeepHash is consistent across data sets.
Compared to {\it ITQ} scheme which applies a single PCA transform, each output bit for DeepHash is generated by a series of projections. 
The {\it PQ} scheme performs poorly at the low rates in consideration, as large blocks of the global descriptor are quantized with a small number of centroids, as previously observed in~\cite{BPBC}.
{\it LSH} performs poorly at low rates, but catches up given enough bits. 

We observe a consistent improvement using Siamese fine-tuning, which learns more discriminative projections.
The learnt projections generalize well, which is key for diverse retrieval tasks, thus showing the robustness of our proposed method.

\vspace{-0.1em}
\paragraph{Comparing FV-DeepHash and DCNN-DeepHash.}
At a given rate point, DCNN-DeepHash outperforms FV-DeepHash hashes for all data sets except {\it Graphics} as seen by comparing across rows of Figure~\ref{fig:retrieval_small_scale}.
At low rates, DCNN-DeepHash improves performance by more than $10\%$ on some of the small data sets, while the gap increases up to 20$\%$  for the lM experiments. 

The results are data-set dependent. 
DCNN features are able to capture more complex low level features and have a lower starting dimensionality compared to FV. 
However, DCNN features have limited rotation and scale invariance, based on the level of data invariance seen at training time.
FV, on the other hand, aggregate hand-crafted SIFT descriptors from scale and rotation invariant interest points, which results in a scale and rotation invariant representation.
The {\it Graphics} data set has more objects with large variations in scale and rotation compared to the other data sets: this is one of the reasons, why peak performance of FV is higher than DCNN for {\it Graphics}.



\vspace{-0.1em}
\paragraph{Comparison to Uncompressed Descriptors.}
We compare the performance of DeepHash to the uncompressed descriptor in Figure~\ref{fig:retrieval_small_scale}.
We obtain remarkable compression performance - at 256 bits for DCNN hashes, we only observe a marginal drop (a few$\%$) compared to the uncompressed representation for retrieval on a wide range of data sets: a 512 $\times$ compression compared to a floating point representation, and 16 $\times$ compared to a binary representation.
For FV, we can match the performance of the uncompressed descriptor with 1024 bits for {\it Holidays} and {\it UKB}, with a  drop for {\it Graphics} and {\it Oxford}. 
At some rate points, DeepHash performs better than the uncompressed descriptor, which is due to quantization of noise in the uncompressed descriptor.

The instance retrieval hashing problem becomes increasingly difficult as we move towards a 64-bit hash.
At 64 bits, there is a 5-10$\%$ drop in performance compared to 256 bits for DCNN features, while a drop is also observed for FV.
For the million scale experiments, however, we observe a 10-20$\%$ drop in performance at 64 bits compared to 1024 bits for DCNN features.

\vspace{-0.1em}
\paragraph{Future Work.} Improving performance further at very low-rate points like 64 bits for even larger databases is an interesting direction for future work.
Studying mathematical models which relate hash size to performance for varying database size is also an exciting direction to pursue.
Finally, in this work, we learnt compact hashes starting from a pre-trained DCNN model.
Learning the hash directly from pixels in a DCNN framework might lead to further improvements.

\section{Conclusion}

A perfect image hashing scheme would convert a high-dimensional descriptor into a low-dimensional bit representation without losing retrieval performance. 
We believe that DeepHash, which focuses on achieving complex hash functions with deep learning, is a significant step in this direction. 
Our method is focused on a deep network which efficiently utilizes the binary subspace through hashing regularization and further fine-tuning using a Siamese training algorithm. 
Through a rigorous evaluation process, we show that our model performs well across various data sets, regardless of the type of image descriptors used, sparse or dense. 
The marked improvement over existing hashing schemes attests to the importance of regularization, depth and fine-tuning for hashing image descriptors.

\let\oldthebibliography=\thebibliography
  \let\endoldthebibliography=\endthebibliography
  \renewenvironment{thebibliography}[1]{%
    \begin{oldthebibliography}{#1}%
      \setlength{\parskip}{0ex}%
      \setlength{\itemsep}{0ex}%
  }%
  {%
    \end{oldthebibliography}%
  }

{
\scriptsize{
\bibliography{main}   

\begin{thebibliography}{10}
\providecommand{\url}[1]{#1}
\csname url@samestyle\endcsname
\providecommand{\newblock}{\relax}
\providecommand{\bibinfo}[2]{#2}
\providecommand{\BIBentrySTDinterwordspacing}{\spaceskip=0pt\relax}
\providecommand{\BIBentryALTinterwordstretchfactor}{4}
\providecommand{\BIBentryALTinterwordspacing}{\spaceskip=\fontdimen2\font plus
\BIBentryALTinterwordstretchfactor\fontdimen3\font minus
  \fontdimen4\font\relax}
\providecommand{\BIBforeignlanguage}[2]{{%
\expandafter\ifx\csname l@#1\endcsname\relax
\typeout{** WARNING: IEEEtran.bst: No hyphenation pattern has been}%
\typeout{** loaded for the language `#1'. Using the pattern for}%
\typeout{** the default language instead.}%
\else
\language=\csname l@#1\endcsname
\fi
#2}}
\providecommand{\BIBdecl}{\relax}
\BIBdecl

\bibitem{Perronnin_CVPR_10}
F.~Perronnin, Y.~Liu, J.~Sanchez, and H.~Poirier, ``{Large-scale Image
  Retrieval with Compressed Fisher Vectors},'' in \emph{Proceedings of IEEE
  Conference on Computer Vision and Pattern Recognition (CVPR)}, San Francisco,
  CA, USA, June 2010, pp. 3384--3391.

\bibitem{AlexNet}
A.~Krizhevsky, I.~Sutskever, and G.~E. Hinton, ``{Imagenet classification with
  deep convolutional neural networks},'' in \emph{Advances in Neural
  Information Processing Systems (NIPS)}, 2012.

\bibitem{Yandex}
A.~Babenko, A.~Slesarev, A.~Chigorin, and V.~Lempitsky, ``{Neural Codes for
  Image Retrieval},'' in \emph{Proceedings of European Conference on Computer
  Vision (ECCV)}, 2014.

\bibitem{VeryDeepNeuralNets}
\BIBentryALTinterwordspacing
K.~Simonyan and A.~Zisserman, ``Very deep convolutional networks for
  large-scale image recognition,'' \emph{CoRR}, vol. abs/1409.1556, 2014.
  [Online]. Available: \url{http://arxiv.org/abs/1409.1556}
\BIBentrySTDinterwordspacing

\bibitem{deepface}
``{DeepFace: Closing the Gap to Human-Level Performance in Face
  Verification},'' in \emph{Conference on Computer Vision and Pattern
  Recognition (CVPR)}.

\bibitem{deepid}
Y.~Sun, X.~Wang, and X.~Tang, ``Deep learning face representation from
  predicting 10,000 classes,'' June 2014.

\bibitem{deeppose}
A.~Toshev and C.~Szegedy, ``Deeppose: Human pose estimation via deep neural
  networks,'' June 2014.

\bibitem{HintonRBM}
R.~Salakhutdinov, A.~Mnih, and G.~Hinton, ``{Restricted Boltzmann Machines for
  Collaborative Filtering},'' in \emph{Proceedings of the 24th International
  Conference on Machine Learning}, ser. ICML '07.\hskip 1em plus 0.5em minus
  0.4em\relax New York, NY, USA: ACM, 2007, pp. 791--798.

\bibitem{ITQ}
Y.~Gong and S.~Lazebnik, ``{Iterative Quantization: A Procrustean Approach to
  Learning Binary Codes},'' in \emph{Proceedings of the 2011 IEEE Conference on
  Computer Vision and Pattern Recognition}, ser. CVPR '11, Washington, DC, USA,
  2011, pp. 817--824.

\bibitem{SpectralHashing}
Y.~Weiss, A.~Torralba, and R.~Fergus, ``{Spectral Hashing},'' in
  \emph{Proceedings of Neural Information Processing Systems (NIPS)},
  Vancouver, BC, Canada, December 2008, pp. 1753--1760.

\bibitem{MLH}
M.~Norouzi and D.~Fleet, ``{Minimal Loss Hashing for Compact Binary Codes},''
  in \emph{Proceedings of ICML}, 2011, pp. 353--360.

\bibitem{KLSH}
B.~Kulis and K.~Grauman, ``{Kernelized Locality-Sensitive Hashing for Scalable
  Image Search},'' in \emph{Proceedings of IEEE International Conference on
  Computer Vision (ICCV}, 2009.

\bibitem{RSH}
J.~Wang, W.~Liu, A.~X. Sun, and Y.~Jiang, ``{Learning Hash Codes with Listwise
  Supervision},'' in \emph{Proceedings of IEEE International Conference on
  Computer Vision, {ICCV} 2013, Sydney, Australia, December 1-8, 2013}, 2013,
  pp. 3032--3039.

\bibitem{CGH}
X.~Li, G.~Lin, C.~Shen, A.~van~den Hengel, and A.~R. Dick, ``{Learning Hash
  Functions Using Column Generation.}'' in \emph{Proceedings of ICML}, vol.~28,
  2013, pp. 142--150.

\bibitem{SemiSupervisedHashing}
J.~Wang, S.~Kumar, and S.-F. Chang, ``{Semi-Supervised Hashing for Scalable
  Image Retrieval},'' in \emph{Proceedings of IEEE Computer Society Conference
  on Computer Vision and Pattern Recognition (CVPR)}, San Francisco, USA, June
  2010.

\bibitem{SphericalHashing}
Y.~Lee, ``{Spherical Hashing},'' in \emph{Proceedings of the 2012 IEEE
  Conference on Computer Vision and Pattern Recognition (CVPR)}.\hskip 1em plus
  0.5em minus 0.4em\relax Washington, DC, USA: IEEE Computer Society, 2012, pp.
  2957--2964.

\bibitem{KristenHashingSurvey}
K.~Grauman and R.~Fergus, ``{Learning Binary Hash Codes for Large-Scale Image
  Search},'' in \emph{Machine Learning for Computer Vision}, 2013, vol. 411,
  pp. 49--87.

\bibitem{SKH}
W.~Liu, J.~Wang, R.~Ji, Y.-G. Jiang, and S.-F. Chang, ``{Supervised Hashing
  with Kernels},'' in \emph{Computer Vision and Pattern Recognition (CVPR)},
  2012.

\bibitem{SmallCodes}
A.~Torralba, R.~Fergus, and Y.~Weiss, ``{Small Codes and Large Image Databases
  for Recognition},'' in \emph{Proceedings of IEEE Conference on Computer
  Vision and Pattern Recognition (CVPR)}, Anchorage, Alaska, June 2008, pp.
  1--8.

\bibitem{SIFTSurvey}
V.~Chandrasekhar, M.~Makar, G.~Takacs, D.~Chen, S.~S. Tsai, N.~M. Cheung,
  R.~Grzeszczuk, Y.~Reznik, and B.~Girod, ``{Survey of SIFT Compression
  Schemes},'' in \emph{Proceedings of International Mobile Multimedia Workshop
  (IMMW), IEEE International Conference on Pattern Recognition (ICPR)},
  Istanbul, Turkey, August 2010.

\bibitem{CHoG}
V.~Chandrasekhar, G.~Takacs, D.~M. Chen, S.~S. Tsai, R.~Grzeszczuk, and
  B.~Girod, ``{CHoG: Compressed Histogram of Gradients - A Low Bit Rate Feature
  Descriptor},'' in \emph{Proceedings of IEEE Conference on Computer Vision and
  Pattern Recognition (CVPR)}, Miami, Florida, June 2009, pp. 2504--2511.

\bibitem{RandomProjections}
C.~Yeo, P.~Ahammad, and K.~Ramchandran, ``{Rate-efficient Visual
  Correspondences using Random Projections},'' in \emph{Proceedings of IEEE
  International Conference on Image Processing (ICIP)}, San Diego, California,
  October 2008, pp. 217--220.

\bibitem{BPBC}
Y.~Gong, S.~Kumar, H.~Rowley, and S.~Lazebnik, ``{Learning Binary Codes for
  High-Dimensional Data Using Bilinear Projections.}'' in \emph{Proceedings of
  CVPR}, 2013, pp. 484--491.

\bibitem{PQFisher}
H.~J{\'e}gou, F.~Perronnin, M.~Douze, J.~S{\'a}nchez, P.~P{\'e}rez, and
  C.~Schmid, ``{Aggregating local image descriptors into compact codes},''
  \emph{{IEEE Transactions on Pattern Analysis and Machine Intelligence}},
  vol.~34, no.~9, pp. 1704--1716, Sep. 2012.

\bibitem{Caffe}
Y.~Jia, E.~Shelhamer, J.~Donahue, S.~Karayev, J.~Long, R.~Girshick,
  S.~Guadarrama, and T.~Darrell, ``Caffe: Convolutional architecture for fast
  feature embedding,'' \emph{arXiv preprint arXiv:1408.5093}, 2014.

\bibitem{HintonDBN}
G.~E. Hinton, S.~Osindero, and Y.-W. Teh, ``A fast learning algorithm for deep
  belief networks,'' \emph{Neural Computation}, vol.~18, no.~7, pp. 1527--1554,
  2006.

\bibitem{siamesenetwork}
J.~Bromley, I.~Guyon, Y.~Lecun, E.~Säckinger, and R.~Shah, ``{Signature
  Verification using a "Siamese" Time Delay Neural Network},'' in
  \emph{Advances in Neural Information Processing Systems (NIPS)}, 1994.

\bibitem{agrawal14analyzing}
P.~Agrawal, R.~Girshick, and J.~Malik, ``Analyzing the performance of
  multilayer neural networks for object recognition,'' in \emph{Proceedings of
  the European Conference on Computer Vision ({ECCV})}, 2014.

\bibitem{RBMPracticalGuide}
G.~Hinton, ``{A Practical Guide to Training Restricted Boltzmann Machines},''
  in \emph{Neural Networks: Tricks of the Trade}, ser. LNCS, vol. 7700.\hskip
  1em plus 0.5em minus 0.4em\relax Springer Berlin Heidelberg, 2012, pp.
  599--619.

\bibitem{hintonCD}
G.~E. Hinton, ``Training products of experts by minimizing contrastive
  divergence,'' \emph{Neural Computation}, vol.~14, no.~8, p.
  1771{\textendash}1800, 2002.

\bibitem{honglakSparsity}
H.~Lee, C.~Ekanadham, and A.~Ng, ``Sparse deep belief net model for visual area
  {V}2,'' in \emph{Advances in Neural Information Processing Systems (NIPS)},
  2008, pp. 873--880.

\bibitem{hintonSparsity}
V.~Nair and G.~Hinton, ``{3{D} Object Recognition with Deep Belief Nets},'' in
  \emph{Advances in Neural Information Processing Systems (NIPS)}, 2009, pp.
  1339--1347.

\bibitem{hanlinSparsity}
H.~Goh, N.~Thome, M.~Cord, and J.-H. Lim, ``Unsupervised and supervised visual
  codes with restricted {B}oltzmann machines,'' in \emph{European Conference on
  Computer Vision (ECCV)}, 2012.

\bibitem{Siamese}
R.~Hadsell, S.~Chopra, and Y.~LeCun, ``{Dimensionality Reduction by Learning an
  Invariant Mapping},'' in \emph{Proceedings of IEEE Conference on Computer
  Vision and Pattern Recognition}, vol.~2, 2006, pp. 1735--1742.

\bibitem{chopra2005}
S.~Chopra, R.~Hadsell, and Y.~Lecun, ``{Learning a similarity metric
  discriminatively, with application to face verification},'' in
  \emph{Proceedings of IEEE Conference on Computer Vision and Pattern
  Recognition (CVPR)}.\hskip 1em plus 0.5em minus 0.4em\relax IEEE Press, 2005,
  pp. 539--546.

\bibitem{Lowe04}
D.~Lowe, ``{Distinctive Image Features from Scale-invariant Keypoints},''
  \emph{International Journal of Computer Vision}, vol.~60, no.~2, pp. 91--110,
  November 2004.

\bibitem{Jegou_CVPR_10}
H.~J\'{e}gou, M.~Douze, C.~Schmid, and P.~Perez, ``{Aggregating Local
  Descriptors into a Compact Image Representation},'' in \emph{Proceedings of
  IEEE Conference on Computer Vision and Pattern Recognition (CVPR)}, San
  Francisco, CA, USA, June 2010, pp. 3304--3311.

\bibitem{SFCV}
J.~Lin, L.-Y. Duan, T.~Huang, and W.~Gao, ``{Robust Fisher Codes for Large
  Scale Image Retrieval},'' in \emph{Proceedings of International Conference on
  Acoustics and Signal Processing (ICASSP)}, 2013.

\bibitem{DengImagenet}
J.~Deng, W.~Dong, R.~Socher, L.-J. Li, K.~Li, and L.~Fei-Fei, ``Imagenet: A
  large-scale hierarchical image database,'' in \emph{IEEE Conference on
  Computer Vision and Pattern Recognition (CVPR)}, 2009.

\bibitem{Philbin07}
J.~Philbin, O.~Chum, M.~Isard, J.~Sivic, and A.~Zisserman, ``{Object Retrieval
  with Large Vocabularies and Fast Spatial Matching},'' in \emph{Proceedings of
  CVPR}, Minneapolis, Minnesota, June 2007, pp. 1--8.

\bibitem{Jegou08}
H.~J\'{e}gou, M.~Douze, and C.~Schmid, ``{Hamming Embedding and Weak Geometric
  Consistency for Large Scale Image Search},'' in \emph{Proceedings of European
  Conference on Computer Vision (ECCV)}, Berlin, Heidelberg, October 2008, pp.
  304--317.

\bibitem{SVMSDataSet}
V.~Chandrasekhar, D.M.Chen, S.S.Tsai, N.M.Cheung, H.Chen, G.Takacs, Y.Reznik,
  R.Vedantham, R.Grzeszczuk, J.Back, and B.Girod, ``{Stanford Mobile Visual
  Search Data Set},'' in \emph{{Proceedings of ACM Multimedia Systems
  Conference (MMSys), San Jose, California, February 2011}}.

\bibitem{MPEGDataset2}
\emph{{MPEG CDVS (Compact Descriptors for Visual Search) Benchmark. Stanford
  Digital Repository}}, \url{http://purl.stanford.edu/qy869qz5226}.

\bibitem{Nister06}
D.~Nist\'er and H.~Stew\'enius, ``{Scalable Recognition with a Vocabulary
  Tree},'' in \emph{Proceedings of IEEE Conference on Computer Vision and
  Pattern Recognition (CVPR)}, New York, USA, June 2006, pp. 2161--2168.

\bibitem{mirflickr}
B.~T. Mark J.~Huiskes and M.~S. Lew, ``{New Trends and Ideas in Visual Concept
  Detection: The MIR Flickr Retrieval Evaluation Initiative},'' in
  \emph{Proceedings of the 2010 ACM International Conference on Multimedia
  Information Retrieval}, New York, NY, USA, 2010, pp. 527--536.

\bibitem{DavidChenThesis}
D.~M. Chen, ``{Memory-Efficient Image Databases for Mobile Visual Search},''
  \emph{Ph.D. thesis, Department of Electrical Engineering, Stanford
  University}, April 2014.

\bibitem{REVV1}
D.~M. Chen, S.~S. Tsai, V.~Chandrasekhar, G.~Takacs, R.~Vedantham,
  R.~Grzeszczuk, and B.~Girod, ``{Residual Enhanced Visual Vector as a Compact
  Signature for Mobile Visual Search},'' in \emph{Signal Processing, Elsevier,
  In Press}, June 2012.

\bibitem{Fischler81}
M.~A. Fischler and R.~C. Bolles, ``{Random Sample Consensus: A Paradigm for
  Model Fitting with Applications to Image Analysis and Automated
  Cartography},'' \emph{Communications of ACM}, vol.~24, no.~6, pp. 381--395,
  June 1981.

\end{thebibliography}
\bibliographystyle{IEEEtran}
}}

\end{document}